\documentclass[conference]{IEEEtran}
\pdfoutput=1
\usepackage{amsmath,amssymb,amsfonts, amsthm}
\usepackage{graphicx}
\usepackage{textcomp}
\usepackage{xcolor}
\def\BibTeX{{\rm B\kern-.05em{\sc i\kern-.025em b}\kern-.08em
    T\kern-.1667em\lower.7ex\hbox{E}\kern-.125emX}}
\usepackage[colorlinks=false, hidelinks]{hyperref}
\usepackage[capitalize]{cleveref}
\usepackage[detect-all,per-mode=symbol]{siunitx}
\usepackage{booktabs}
\newcolumntype{M}{>{$}l<{$}} 
\usepackage{multirow}
\usepackage{pgfplots}
\usepackage[protrusion=true,expansion=true]{microtype} 
\usepackage{algorithm}
\usepackage{algorithmic}
\usepackage{todonotes}
\usepackage[nolist,printonlyused]{acronym}
\usepackage{standalone}
\usepackage{enumitem}
\usepackage[font=small,labelfont=bf]{caption}

\newcommand{\gf}[1]{\textcolor{cyan}{{#1}}}
\newcommand{\sam}[1]{\textcolor{brown}{{#1}}}
\newcommand{\ps}[1]{\textcolor{blue}{{#1}}}

\usepgfplotslibrary{groupplots}
\usetikzlibrary{positioning}
\pgfplotsset{compat=newest}
\pgfplotsset{every axis legend/.append style={%
cells={anchor=west}}
}
\pgfplotsset{every axis/.append style={
                    label style={font=\footnotesize},
					tick label style={font=\footnotesize},
					legend style={font=\footnotesize}
                    }}

\usepackage[style=ieee, maxcitenames=2, mincitenames=1, backend=biber]{biblatex}
\AtEveryBibitem{
 	\clearfield{url}
 	\clearfield{doi}
\ifentrytype{inproceedings}{
 	\clearlist{address}
 	\clearlist{publisher}
 	\clearname{editor}
 	\clearlist{organization}
 	\clearfield{pages}
 	\clearlist{location}
	\clearfield{volume}
	\clearfield{series}
	\clearfield{eprint}
	\clearfield{eprinttype}
 }{}
 }
\AtBeginBibliography{\footnotesize}

\usepackage{float}

\usepackage{comment}

\newcommand{\addappendix}{}     

\begin{document}

\title{Offline Reinforcement Learning and Sequence Modeling for Downlink Link Adaptation}
\author{
    Samuele Peri\\
    \textit{Politecnico di Milano} \\
  \and
    Alessio Russo\\
    \textit{Ericsson AB}
    \and
    Gabor Fodor\\
    \textit{Ericsson AB}
    \and
    Pablo Soldati\\
    \textit{Ericsson AB}
}

\maketitle


\begin{abstract}

\Ac{LA} is an essential function in modern wireless communication systems that dynamically adjusts the transmission rate of a communication link to match time- and frequency-varying radio link conditions.
However, factors such as user mobility, fast fading, imperfect channel quality information, and aging of measurements make the modeling of LA challenging. 
%
To bypass the need for explicit modeling, recent research has introduced online \ac{RL} approaches as an alternative to the more commonly used rule-based algorithms.
Yet, RL-based approaches face deployment challenges, as training in live networks can potentially degrade real-time performance.
To address this challenge, this paper considers offline RL as~a candidate to learn LA policies with minimal effects~on the network operation. 
We propose three LA designs based on batch-constrained deep Q-learning, conservative Q-learning, and decision transformer. 
Our results show that offline~RL algorithms can match the performance of state-of-the-art online RL methods when data is collected with a proper behavioral policy.
\end{abstract}

\acresetall 


\begin{IEEEkeywords}%
  Offline reinforcement learning; decision transformer; link adaptation; radio access networks.%
\end{IEEEkeywords}


\acrodefplural{MDP}[MDPs]{Markov decision processes}
\acrodefplural{RTG}[RTGs]{returns-to-go}

\begin{acronym}
\acro{ACK}{acknowledgement}
\acro{ARQ}{automatic repeat request}
\acro{BE}{baseline encoding}
\acro{BC}{behavioral cloning}
  \acro{BCQ}{batch-constrained deep Q-learning}
  \acro{CCTR}{Channel-Conditioned Target Return}
  \acro{CNN}{convolutional neural network}
  \acro{CQL}{conservative Q-learning}
  \acro{CT}{continuous time}
  \acro{CV}{computer vision}
  \acro{DAVG}{discounted average}
  \acro{DP}{dynamic programming}
  \acro{DQN}{deep Q-network}
  \acro{DT}{decision transformer}
  \acro{eMBB}{enhanced mobile broadband}
  \acro{ILLA}{inner-loop link adaptation}
  \acro{gNB}{next generation NodeB}
  \acro{gNB-CU}{gNB centralized unit}
  \acro{gNB-DU}{gNB distributed unit}
  \acro{GPI}{generalized policy iteration}
  \acro{LA}{link adaptation}
  \acro{LSTM}{long short-term memory}
  \acro{LT}{learnable time}
  \acro{MC}{Monte Carlo}
  \acro{MDP}{Markov decision process}
  \acro{ML}{machine learning}
  \acro{MLP}{multi-layer perceptron}
  \acro{mMIMO}{massive multiple input multiple output}
  \acro{mMTC}{massive machine-type communications}
  \acro{NACK}{negative acknowledgement}
  \acro{NLP}{natural language processing}
  \acro{OLLA}{outer-loop link adaptation}
  \acro{PE}{positional encoding}
  \acro{PHY}{physical layer}
  \acro{RL}{reinforcement learning}
  \acro{RLC}{radio link control}
  \acro{RNN}{recurrent neural network}
  \acro{RTGs}{returns-to-go}
  \acro{RTG}{return-to-go}
  \acro{RvS}{reinforcement learning via supervised learning}
  \acro{SACo}{state-action coverage}
  \acro{SADCo}{state-action density coverage}
  \acro{SE}{spectral efficiency}
  \acro{TBS}{transport block size}
  \acro{TD}{temporal difference}
  \acro{TQ}{relative trajectory quality}
  \acro{URLLC}{ultra-reliable low-latency communications}
  \acro{VAE}{variational auto-encoder}
  \acro{2G}{Second Generation}
  \acro{3G}{3$^\text{rd}$~Generation}
  \acro{3GPP}{3$^\text{rd}$~Generation Partnership Project}
  \acro{4G}{4$^\text{th}$~Generation}
  \acro{5G}{5$^\text{th}$~Generation}
  \acro{AA}{Antenna Array}
  \acro{AC}{Admission Control}
  \acro{AD}{Attack-Decay}
  \acro{ADSL}{Asymmetric Digital Subscriber Line}
	\acro{AHW}{Alternate Hop-and-Wait}
  \acro{AMC}{Adaptive Modulation and Coding}
  \acro{AoA}{angle of arrival}
	\acro{AP}{Access Point}
  \acro{APA}{Adaptive Power Allocation}
  \acro{AR}{autoregressive}
  \acro{ARMA}{Autoregressive Moving Average}
  \acro{ATES}{Adaptive Throughput-based Efficiency-Satisfaction Trade-Off}
  \acro{AWGN}{additive white Gaussian noise}
  \acro{BB}{Branch and Bound}
  \acro{BD}{Block Diagonalization}
  \acro{BER}{bit error rate}
  \acro{BF}{Best Fit}
  \acro{BLER}{block error rate}
  \acro{BPC}{Binary power control}
  \acro{BPSK}{Binary Phase-Shift Keying}
  \acro{BPA}{Best \ac{PDPR} Algorithm}
  \acro{BRA}{Balanced Random Allocation}
  \acro{BCRB}{Bayesian Cram\'{e}r-Rao Bound}
  \acro{BS}{base station}
  \acro{CAP}{Combinatorial Allocation Problem}
  \acro{CAPEX}{Capital Expenditure}
  \acro{CBF}{Coordinated Beamforming}
  \acro{CBR}{Constant Bit Rate}
  \acro{CBS}{Class Based Scheduling}
  \acro{CC}{Congestion Control}
  \acro{CDF}{cumulative distribution function}
  \acro{CDMA}{Code-Division Multiple Access}
  \acro{CL}{Closed Loop}
  \acro{CLPC}{Closed Loop Power Control}
  \acro{CNR}{Channel-to-Noise Ratio}
  \acro{CPA}{Cellular Protection Algorithm}
  \acro{CPICH}{Common Pilot Channel}
  \acro{CoMP}{Coordinated Multi-Point}
  \acro{CQI}{channel quality indicator}
  \acro{CRB}{Cram\'{e}r-Rao Bound}
  \acro{CRM}{Constrained Rate Maximization}
	\acro{CRN}{Cognitive Radio Network}
  \acro{CS}{Coordinated Scheduling}
  \acro{CSI}{channel state information}
  \acro{CSIR}{channel state information at the receiver}
  \acro{CSIT}{channel state information at the transmitter}
  \acro{CUE}{cellular user equipment}
  \acro{D2D}{device-to-device}
  \acro{DCA}{Dynamic Channel Allocation}
  \acro{DE}{Differential Evolution}
  \acro{DFT}{Discrete Fourier Transform}
  \acro{DIST}{Distance}
  \acro{DL}{downlink}
  \acro{DMA}{Double Moving Average}
	\acro{DMRS}{demodulation reference signal}
  \acro{D2DM}{D2D Mode}
  \acro{DMS}{D2D Mode Selection}
  \acro{DPC}{Dirty Paper Coding}
  \acro{DRA}{Dynamic Resource Assignment}
  \acro{DSA}{Dynamic Spectrum Access}
  \acro{DSM}{Delay-based Satisfaction Maximization}
  \acro{ECC}{Electronic Communications Committee}
  \acro{EFLC}{Error Feedback Based Load Control}
  \acro{EI}{Efficiency Indicator}
  \acro{eNB}{Evolved Node B}
  \acro{EPA}{Equal Power Allocation}
  \acro{EPC}{Evolved Packet Core}
  \acro{EPS}{Evolved Packet System}
  \acro{ESPRIT}{estimation of signal parameters via rotational invariance}
  \acro{E-UTRAN}{Evolved Universal Terrestrial Radio Access Network}
  \acro{ES}{Exhaustive Search}
  \acro{FDD}{frequency division duplexing}
  \acro{FDM}{Frequency Division Multiplexing}
  \acro{FER}{Frame Erasure Rate}
  \acro{FF}{Fast Fading}
  \acro{FIM}{Fisher information matrix}
  \acro{FSB}{Fixed Switched Beamforming}
  \acro{FST}{Fixed SNR Target}
  \acro{FTP}{File Transfer Protocol}
  \acro{GA}{Genetic Algorithm}
  \acro{GBR}{Guaranteed Bit Rate}
  \acro{GLR}{Gain to Leakage Ratio}
  \acro{GOS}{Generated Orthogonal Sequence}
  \acro{GPL}{GNU General Public License}
  \acro{GRP}{Grouping}
  \acro{HARQ}{hybrid automatic repeat request}
  \acro{HMS}{Harmonic Mode Selection}
  \acro{HOL}{Head Of Line}
  \acro{HSDPA}{High-Speed Downlink Packet Access}
  \acro{HSPA}{High Speed Packet Access}
  \acro{HTTP}{HyperText Transfer Protocol}
  \acro{ICMP}{Internet Control Message Protocol}
  \acro{ICI}{Intercell Interference}
  \acro{ID}{Identification}
  \acro{ISAC}{integrated sensing and communication}
  \acro{IEEE}{Institute of Electrical and Electronics Engineers}
  \acro{IETF}{Internet Engineering Task Force}
  \acro{ILP}{Integer Linear Program}
  \acro{JRAPAP}{Joint RB Assignment and Power Allocation Problem}
  \acro{UID}{Unique Identification}
  \acro{IID}{Independent and Identically Distributed}
  \acro{IIR}{Infinite Impulse Response}
  \acro{ILP}{Integer Linear Problem}
  \acro{IMT}{International Mobile Telecommunications}
  \acro{INV}{Inverted Norm-based Grouping}
  \acro{IoT}{Internet of Things}
  \acro{IP}{Internet Protocol}
  \acro{IPv6}{Internet Protocol Version 6}
  \acro{ISD}{Inter-Site Distance}
  \acro{ISI}{Inter Symbol Interference}
  \acro{ITU}{International Telecommunication Union}
  \acro{JOAS}{Joint Opportunistic Assignment and Scheduling}
  \acro{JOS}{Joint Opportunistic Scheduling}
  \acro{JP}{Joint Processing}
  \acro{JS}{Jump-Stay}
  \acro{KKT}{Karush-Kuhn-Tucker}
  \acro{L3}{Layer-3}
  \acro{LAC}{Link Admission Control}
  \acro{LC}{Load Control}
  \acro{LOS}{Line of Sight}
  \acro{LP}{Linear Programming}
  \acro{LS}{least squares}
  \acro{LTE}{Long Term Evolution}
  \acro{LTE-A}{LTE-Advanced}
  \acro{LTE-Advanced}{Long Term Evolution Advanced}
  \acro{M2M}{Machine-to-Machine}
  \acro{MAC}{Medium Access Control}
  \acro{MANET}{Mobile Ad hoc Network}
  \acro{MCS}{modulation and coding scheme}
  \acro{MDB}{Measured Delay Based}
  \acro{MDI}{Minimum D2D Interference}
  \acro{MF}{Matched Filter}
  \acro{MG}{Maximum Gain}
  \acro{MH}{Multi-Hop}
  \acro{MIMO}{multiple input multiple output}
  \acro{MINLP}{Mixed Integer Nonlinear Programming}
  \acro{MIP}{Mixed Integer Programming}
  \acro{MISO}{Multiple Input Single Output}
  \acro{MLE}{maximum likelihood estimator}
  \acro{MLWDF}{Modified Largest Weighted Delay First}
  \acro{MME}{Mobility Management Entity}
  \acro{MMSE}{minimum mean squared error}
  \acro{MOS}{Mean Opinion Score}
  \acro{MPF}{Multicarrier Proportional Fair}
  \acro{MRA}{Maximum Rate Allocation}
  \acro{MR}{Maximum Rate}
  \acro{MRC}{Maximum Ratio Combining}
  \acro{MRT}{Maximum Ratio Transmission}
  \acro{MRUS}{Maximum Rate with User Satisfaction}
  \acro{MS}{mobile station}
  \acro{MSE}{mean squared error}
  \acro{MSI}{Multi-Stream Interference}
  \acro{MTC}{Machine-Type Communication}
  \acro{MTSI}{Multimedia Telephony Services over IMS}
  \acro{MTSM}{Modified Throughput-based Satisfaction Maximization}
  \acro{MU-MIMO}{multiuser multiple input multiple output}
  \acro{MU}{multi-user}
  \acro{MUSIC}{multiple signal classification}
  \acro{NAS}{Non-Access Stratum}
  \acro{NB}{Node B}
  \acro{NE}{Nash equilibrium}
  \acro{NCL}{Neighbor Cell List}
  \acro{NLOS}{Non-Line of Sight}
  \acro{NMSE}{Normalized Mean Square Error}
  \acro{NORM}{Normalized Projection-based Grouping}
  \acro{NP}{Non-Polynomial Time}
  \acro{NR}{New Radio}
  \acro{NRT}{Non-Real Time}
  \acro{NSPS}{National Security and Public Safety Services}
  \acro{O2I}{Outdoor to Indoor}
  \acro{OFDMA}{orthogonal frequency division multiple access}
  \acro{OFDM}{orthogonal frequency division multiplexing}
  \acro{OFPC}{Open Loop with Fractional Path Loss Compensation}
	\acro{O2I}{Outdoor-to-Indoor}
  \acro{OL}{Open Loop}
  \acro{OLPC}{Open-Loop Power Control}
  \acro{OL-PC}{Open-Loop Power Control}
  \acro{OPEX}{Operational Expenditure}
  \acro{ORB}{Orthogonal Random Beamforming}
  \acro{JO-PF}{Joint Opportunistic Proportional Fair}
  \acro{OSI}{Open Systems Interconnection}
  \acro{PAIR}{D2D Pair Gain-based Grouping}
  \acro{PAPR}{Peak-to-Average Power Ratio}
  \acro{P2P}{Peer-to-Peer}
  \acro{PC}{Power Control}
  \acro{PCI}{Physical Cell ID}
  \acro{PDF}{Probability Density Function}
  \acro{PDPR}{pilot-to-data power ratio}
  \acro{PER}{Packet Error Rate}
  \acro{PF}{Proportional Fair}
  \acro{P-GW}{Packet Data Network Gateway}
  \acro{PL}{Pathloss}
  \acro{PPR}{pilot power ratio}
  \acro{PRB}{physical resource block}
  \acro{PROJ}{Projection-based Grouping}
  \acro{ProSe}{Proximity Services}
  \acro{PS}{Packet Scheduling}
  \acro{PSAM}{pilot symbol assisted modulation}
  \acro{PSO}{Particle Swarm Optimization}
  \acro{PZF}{Projected Zero-Forcing}
  \acro{QAM}{Quadrature Amplitude Modulation}
  \acro{QoS}{Quality of Service}
  \acro{QPSK}{Quadri-Phase Shift Keying}
  \acro{RAISES}{Reallocation-based Assignment for Improved Spectral Efficiency and Satisfaction}
  \acro{RAN}{radio access network}
  \acro{RAT}{Radio Access Technology}
  \acro{RATE}{Rate-based}
  \acro{RB}{resource block}
  \acro{RBG}{Resource block broup}
  \acro{REF}{Reference Grouping}
  \acro{RM}{Rate Maximization}
  \acro{RNC}{Radio Network Controller}
  \acro{RND}{Random Grouping}
  \acro{RRA}{Radio Resource Allocation}
  \acro{RRM}{radio resource management}
  \acro{RSCP}{Received Signal Code Power}
  \acro{RSRP}{Reference Signal Receive Power}
  \acro{RSRQ}{Reference Signal Receive Quality}
  \acro{RR}{Round Robin}
  \acro{RRC}{Radio Resource Control}
  \acro{RSSI}{Received Signal Strength Indicator}
  \acro{RT}{Real Time}
  \acro{RU}{Resource Unit}
  \acro{RUNE}{RUdimentary Network Emulator}
  \acro{RV}{Random Variable}
  \acro{SAC}{Session Admission Control}
  \acro{SCM}{Spatial Channel Model}
  \acro{SC-FDMA}{Single Carrier - Frequency Division Multiple Access}
  \acro{SD}{Soft Dropping}
  \acro{S-D}{Source-Destination}
  \acro{SDPC}{Soft Dropping Power Control}
  \acro{SDMA}{Space-Division Multiple Access}
  \acro{SER}{Symbol Error Rate}
  \acro{SES}{Simple Exponential Smoothing}
  \acro{S-GW}{Serving Gateway}
  \acro{SINR}{signal-to-interference-plus-noise ratio}
  \acro{SI}{Satisfaction Indicator}
  \acro{SIP}{Session Initiation Protocol}
  \acro{SISO}{single input single output}
  \acro{SIMO}{Single Input Multiple Output}
  \acro{SIR}{signal-to-interference ratio}
  \acro{SLNR}{Signal-to-Leakage-plus-Noise Ratio}
  \acro{SMA}{Simple Moving Average}
  \acro{SNR}{signal-to-noise ratio}
  \acro{SORA}{Satisfaction Oriented Resource Allocation}
  \acro{SORA-NRT}{Satisfaction-Oriented Resource Allocation for Non-Real Time Services}
  \acro{SORA-RT}{Satisfaction-Oriented Resource Allocation for Real Time Services}
  \acro{SPF}{Single-Carrier Proportional Fair}
  \acro{SRA}{Sequential Removal Algorithm}
  \acro{SRS}{Sounding Reference Signal}
  \acro{SSB}{synchronisation signal block}
  \acro{SU-MIMO}{single-user multiple input multiple output}
  \acro{SU}{Single-User}
  \acro{SVD}{Singular Value Decomposition}
  \acro{TCP}{Transmission Control Protocol}
  \acro{TDD}{time division duplexing}
  \acro{TDMA}{Time Division Multiple Access}
  \acro{TETRA}{Terrestrial Trunked Radio}
  \acro{TP}{Transmit Power}
  \acro{TPC}{Transmit Power Control}
  \acro{TTI}{transmission time interval}
  \acro{TTR}{Time-To-Rendezvous}
  \acro{TSM}{Throughput-based Satisfaction Maximization}
  \acro{TU}{Typical Urban}
  \acro{UE}{user equipment}
  \acro{UEPS}{Urgency and Efficiency-based Packet Scheduling}
  \acro{UL}{uplink}
  \acro{UMTS}{Universal Mobile Telecommunications System}
  \acro{URI}{Uniform Resource Identifier}
  \acro{URM}{Unconstrained Rate Maximization}
  \acro{UT}{user terminal}
  \acro{VR}{Virtual Resource}
  \acro{VoIP}{Voice over IP}
  \acro{WAN}{Wireless Access Network}
  \acro{WCDMA}{Wideband Code Division Multiple Access}
  \acro{WF}{Water-filling}
  \acro{WiMAX}{Worldwide Interoperability for Microwave Access}
  \acro{WINNER}{Wireless World Initiative New Radio}
  \acro{WLAN}{Wireless Local Area Network}
  \acro{WMPF}{Weighted Multicarrier Proportional Fair}
  \acro{WPF}{Weighted Proportional Fair}
  \acro{WSN}{Wireless Sensor Network}
  \acro{WWW}{World Wide Web}
  \acro{XIXO}{(Single or Multiple) Input (Single or Multiple) Output}
  \acro{ZF}{zero-forcing}
  \acro{ZMCSCG}{Zero Mean Circularly Symmetric Complex Gaussian}
\end{acronym}

\section{Introduction}\label{sec:introduction}

Modern \acp{RAN} -- such as the \ac{5G} \ac{NR} system -- employ model-based \ac{LA} algorithms to adapt the \ac{MCS} to time-- and frequency-varying channel conditions to maintain high spectral efficiency~\cite{duran_self-optimization_2015, blanquez-casado_eolla_2016, Blanquez-Casado:19, Ramireddy:22, Nicolini:23}. For \ac{DL} communication, \ac{LA} relies on \ac{CSI} reports provided by the \ac{UE} that include a \ac{CQI}, a rank indicator, and a precoder matrix indicator~\cite{3rd_generation_partnership_project_3gpp_technical_2024}.
As \ac{CSI} reports may be inaccurate and rapidly age due to fast channel variations, \ac{OLLA} schemes typically adjust \ac{SINR} estimates inferred from~\ac{CQI} values using \ac{HARQ} feedback~\cite{duran_self-optimization_2015}. Nonetheless, user mobility, fast fading, limited receiver-side information at the transmitter, as well as the heterogeneity of \acp{UE} population, e.g., in terms of receiver capabilities, render \ac{RRM} problems, such as \ac{LA}, complex control tasks to model~\cite{Chen:23}.

Recognizing the challenges faced by model-based approaches, \ac{ML} methods, such as \ac{RL}, have recently gained attention to address \ac{RRM} problems in \acp{RAN}~\cite{calabrese_learning_2018}.
The ability of \ac{RL} to learn directly from experience and to adapt to changes in the environment makes it appealing to replace legacy rule-based \ac{RAN} functions with models learned from data\ifdefined\addappendix~\cite{generalization_2024,gronland2024constrained,mollerstedt2024model}\else~\cite{generalization_2024}\fi. Several works have applied \ac{RL} to \ac{LA}, starting with tabular~\cite{Bruno:14} and bandit~\cite{saxena_reinforcement_2021} methods, adaptations of \ac{OLLA} based on Q-learning~\cite{Chen:23}, and \ac{DQN}~\cite{generalization_2024}.
%
%
However, training \ac{RL} agents in live networks requires exploring the state-action space to learn an optimal policy, at the cost of taking suboptimal actions that may adversely affect the network performance~\cite{russo2024model}.

To address this adverse effect of \ac{RL}, recent works have proposed offline \ac{RL}, which aims to learn an optimal policy from a \emph{static dataset} of transitions produced by a behavioral policy, thus avoiding direct interactions with the environment while training~\cite{levine_offline_2020-1, janner2021offline, schweighofer2022dataset, Kumar:23, lyu2022mildly}. The inability to improve the training dataset during training confines the quality and the behavior of the learned policy to the quality and state-action coverage of the available data. As such, distribution shift and out-of-distribution behavior, that is, a mismatch between the training and testing trajectory distributions, can render the learned policy ineffective.
%
Thus, employing offline \ac{RL} to replace complex real-time \ac{RRM} control functions in \acp{RAN}, such as \ac{LA}, is non-trivial and presents two main challenges: firstly, the ability of an offline \ac{RL} design to achieve, when trained with proper data, optimal performance, e.g., comparable with a policy learned online; secondly, the design of a non-invasive behavioral policy to collect rich data in live networks without affecting performance.

This paper focuses on the first challenge considering offline \ac{RL} for \ac{LA} in \ac{5G} networks. We present three \ac{LA} designs based on \ac{BCQ}~\cite{fujimoto2019off}, \ac{CQL}~\cite{kumar_conservative_2020}, and \ac{DT}~\cite{chen_decision_2021}, with  a trained \ac{DQN} model \cite{mnih_playing_2013} acting as a behavioral policy for data collection. While for \ac{BCQ} and \ac{CQL} we model the \ac{LA} problem as a partially observable \ac{MDP}, \ac{DT} requires a sequence modeling approach, where an \ac{LA} action is inferred using a sequence of historical states, actions and rewards, allowing to optimize long-term rewards, e.g.,~average throughput or spectral efficiency. Our results show that offline \ac{RL} can achieve the performance of online \ac{RL} and outperform practical state-of-the-art \ac{LA} based on \ac{OLLA}  when data is collected with a proper behavioral policy\ifdefined \addappendix \else \footnote{For the full technical report, please refer to~\cite{Peri:24}.}\fi.

The rest of the paper is organized as follows. Section~\ref{sec:background}~introduces preliminary concepts, Section~\ref{sec:problem formulation} formally describes the problem of \ac{LA} while Section~\ref{sec:problem} presents the \ac{BCQ} and \ac{CQL} design for \ac{LA}. Section~\ref{sec:method} frames \ac{LA} as a sequence modeling problem, presenting a \ac{DT}-based design. Section~\ref{sec:experiments} discusses numerical results, while Section~\ref{sec:conclusion} concludes the paper.

\section{Preliminaries}\label{sec:background}
This section reviews some of the key concepts and introduces notation that will be useful in the sequel.

\subsection{Elements of \ac{RL}}

\ac{RL} frames the interaction between an agent and the environment as a \ac{MDP} \cite{Sutton:14,puterman2014markov}, defined by the tuple: $(\mathcal{S, A, P}, r)$. Here, $\mathcal{S}$ is the state-space, $\mathcal{A}$ is the action-space, $\mathcal{P}$ is the transition probability function (also known as transition dynamics), $r:\mathcal{S}\times\mathcal{A}\times\mathcal{S}\to\mathbb{R}$ is the reward function.
In an online setting, at each time-step $t$ the agent chooses an action $a_t$ according to some Markovian policy $\pi:{\cal S}\to {\cal A}$, which is function of the current state $s_t$ of the MDP, and observes the next state $s_{t+1}$, distributed according to $P(\cdot|s_t,a_t)$, and a reward $r_t$. 
For episodic MDPs, with a finite terminal state, the goal of the agent is to find a policy $\pi$ maximizing the expected collected reward $\mathbb{E}^\pi[G_t|s_0\sim \mu]$ for some initial state distribution $\mu$, where
$
    G_t = \sum_{k=0}^T r_{t+k},
$
and $T$ indicates the terminal step. In the context of \textit{continuing tasks}, where $T$ is not naturally defined, or may be infinite, the goal is to maximize the collected discounted reward, defined with a  discount factor $\gamma\in (0,1)$ as $
    \mathbb{E}^\pi[\sum_{k=0}^\infty \gamma^k r_{t+k+1}|s_0\sim \mu]
$.

\subsection{Offline RL}
Differently from the online \ac{RL} setting, in offline \ac{RL} \cite{levine_offline_2020} the agent learns a policy from a previously collected dataset of experiences $\mathcal{D} = \{( s_t^i, a_t^i, r_{t}^i,s_{t+1}^i)_i\}$ (collected using a behavioral policy $\pi_\beta$).
As a consequence, the amount and quality of data, play a crucial role in offline \ac{RL} since the agent is not able to interact with the environment in an online fashion to collect new data. 
The inability to explore the state-action space during training may also result in a distributional shift between the distribution under which the policy $\pi$ is trained and the one on which it will be evaluated \cite{levine_offline_2020}.

To address this issue, some approaches constrain how much the learned policy $\pi$ differs from the behavioral policy~$\pi_\beta$, while other methods penalize the learned $Q$-values. For example, with discrete action spaces, \ac{BCQ} tackles the distributional shift issue by modeling the training data distribution, resulting from data collection with a certain behavioral policy, so that the support of action selection can be restricted to a fixed number of actions sampled from it \cite{fujimoto2019off}.
In contrast, \ac{CQL} \cite{kumar_conservative_2020} uses a \emph{pessimism} principle: it modifies the training objective to minimize, alongside the Bellman error \cite{Sutton:14}, the estimates of the $Q$-values to learn lower-bounded estimates that are~not overly optimistic. A comprehensive survey can be found~in~\cite{levine_offline_2020}.

\subsection{Sequence Modeling and Decision Transformer} 
Sequence modeling involves training auto-regressive models (such as the \ac{DT} \cite{chen_decision_2021}) on sequential data to capture patterns and make future predictions. Since the goal in \ac{RL}~is~to produce a sequence of actions leading to high cumulative rewards, one can re-frame the RL problem in terms of sequence modeling.

The \ac{DT} is a particular auto-regressive architecture used in sequence modeling that exploits the capabilities of GPT2~\cite{radford_language_2019} to perform sequential decision making. For each time step $t$ of a trajectory, \ac{DT}  uses sequences of $3$ embedded tokens: the state $s_t$, the action $a_t$, and the \ac{RTG} $\hat R_t = \sum_{t'=t}^T r_{t'}$. 
At inference time the \ac{RTG} allows to condition the output of the network to generate optimal actions by specifying \textit{future target} returns.
The model is trained by predicting the action $a_t$ at each time-step of the sequence. Overall, this architecture allows to bypass the need for bootstrapping to propagate returns, one of the key destabilizing elements in offline \ac{RL}.

\section{Problem Formulation}\label{sec:problem formulation}
We consider the problem of \ac{LA} in the downlink of multi-cellular \acp{RAN}. The aim of LA is to dynamically adapt the transmission rate of individual packet transmissions to time- and frequency-varying channel conditions and radio interference. In \ac{5G} systems, this is achieved by selecting a set of \ac{MCS} parameters encoded by an integer taking value in the range~$\{0,31\}$,  known as \ac{MCS} index~\cite{3rd_generation_partnership_project_3gpp_technical_2024}. Each \ac{MCS} index corresponds to a modulation order, code rate, and spectral efficiency to be used for transmitting a data packet to the \ac{UE}.

State-of-the-art \ac{LA} algorithms used in real-time \acp{RAN} combine two control loops executed on a sub-millisecond~timescale: an \ac{ILLA} converts the latest \ac{CQI} reported by the \ac{UE} into an initial \ac{SINR} estimate; an \ac{OLLA} scheme adjusts such an estimate (e.g., to compensate inaccuracies or aging of the \ac{CQI}) based on the \ac{HARQ} feedback (i.e., either positive or negative acknowledgments) received from the \ac{UE} for previous packet transmissions. The goal of the \ac{OLLA} is to maximize the link spectral efficiency while adhering to a desired \ac{BLER} target. As variants of this approach are widely adopted in today's communication systems, we consider it as reference for performance comparison in Section~\ref{sec:experiments}.



\section{Link Adaptation as an Offline RL Problem}\label{sec:problem}
This section explores how \ac{LA} can be solved with offline \ac{RL}. We first propose a \ac{MDP} formulation which we not only consider for training offline \ac{RL} agents but, being particularly well-suited for online approaches as well, also for performing data collection. Then, we introduce the minimal modifications that are needed, given the chosen \ac{MDP} formulation, to adapt the \ac{BCQ} and \ac{CQL} algorithms to the \ac{LA} setting.




\begin{figure}
    \centering
    \resizebox{\linewidth}{!}{%
    \begin{tikzpicture}

\tikzstyle{block} = [draw, minimum width=0.75cm, minimum height=0.6cm, align=center]
\tikzstyle{arrow} = [thick,->,>=stealth]
\definecolor{darkgreen}{RGB}{0, 150, 0} 

\node[block, fill=blue!30] (P1) {p$_1$};
\node[block, right=0cm of P1] (E1B2) {};
\node[block, right=0cm of E1B2] (E1B3) {};
\node[block, right=0cm of E1B3] (E1B4) {};
\node[block, right=0cm of E1B4, fill=green!30] (P2) {p$_2$};
\node[block, right=0cm of P2] (E1B5) {};
\node[block, right=0cm of E1B5] (E1B6) {};
\node[block, right=0cm of E1B6] (E1B7) {};
\node[block, right=0cm of E1B7, fill=blue!30] (P3) {p$_1$};
\node[block, right=0cm of P3] (E1B8) {};
\node[block, right=0cm of E1B8] (E1B9) {};
\node[block, right=0cm of E1B9] (E1B10) {};
\node[block, right=0cm of E1B10] (E1B11) {};
\node[block, right=0cm of E1B11] (E1B12) {};
\node[block, right=0cm of E1B12] (E1B13) {};
\node[block, right=0cm of E1B13, fill=pink] (P4) {p$_3$};

\node[block, below=1cm of P1] (E2B1) {};
\node[block, right=0cm of E2B1] (E2B2) {};
\node[block, right=0cm of E2B2] (E2B3) {};
\node[block, right=0cm of E2B3] (E2B4) {};
\node[block, right=0cm of E2B4] (E2B5) {};
\node[block, right=0cm of E2B5] (E2B6) {};
\node[block, fill=blue!30, right=0cm of E2B6] (A1) {};
\node[block, right=0cm of A1] (E2B8) {};
\node[block, right=0cm of E2B8] (E2B9) {};
\node[block, right=0cm of E2B9] (E2B10) {};
\node[block, right=0cm of E2B10] (E2B11) {};
\node[block, fill=green!30, right=0cm of E2B11] (A2) {};
\node[block, right=0cm of A2] (E2B13) {};
\node[block, fill=blue!30, right=0cm of E2B13] (A3) {};
\node[block, right=0cm of A3] (E2B15) {};
\node[block, right=0cm of E2B15] (E2B16) {};

\node[left=0.15cm of P1] (BS) {BS};
\node[left=0.15cm of E2B1] (UE) {UE};

\node[above=0.08cm of A1, shift={(0.6cm, 0)}] (NACK) {NACK};
\node[above=0.08cm of A2, shift={(0.5cm, 0)}] (ACK) {ACK};
\node[above=0.08cm of A3, shift={(0.5cm, 0)}] (ACK2) {ACK};

\draw[arrow, dotted] (A1.north) -- (E1B6.south);
\draw[arrow, dotted] (A2.north) -- (E1B10.south);
\draw[arrow, dotted] (A3.north) -- (E1B12.south);

\draw[arrow] (P1.south) -- (E2B1.north);
\draw[arrow] (P2.south) -- (E2B5.north);
\draw[arrow] (P3.south) -- (E2B9.north);
\draw[arrow] (P4.south) -- (E2B16.north);

\coordinate (E11_start) at (P1.north west |- {0, +0.6}); 
\coordinate (E11_end) at (E1B6.north east |- {0, +0.6});

\coordinate (E21_start) at (P2.north west |- {0, +1.2}); 
\coordinate (E21_end) at (E1B10.north east |- {0, +1.2});

\coordinate (E12_start) at (P3.north west |- {0, +0.6}); 
\coordinate (E12_end) at (E1B12.north east |- {0, +0.6});

\coordinate (E41_start) at (P4.north west |- {0, +0.6}); 
\coordinate (E41_end) at (P4.north east |- {0, +0.6});

\draw[<->] (E11_start) -- (E11_end) node[midway, above] {E1 Step 1 $(s_{1,1}, a_{1,1}, r_{1,1})$};
\draw[<->] (E21_start) -- (E21_end) node[midway, above] {E2 Step 1 $(s_{2,1}, a_{2,1}, r_{2,1}$)};
\draw[<->] (E12_start) -- (E12_end) node[midway, above] {E1 Step 2 $(s_{1,2}, a_{1,2}, r_{1,2})$};
\draw[->] (E41_end) -- (E41_start) node[midway, above] {};

\coordinate (E1_start) at (E2B1.west |- {0, -2.2}); 
\coordinate (E1_end) at (A3.east |- {0, -2.2});

\draw[<->] (E1_start) -- (E1_end) node[midway, below] {Episode 1};

\coordinate (E2_start) at (E2B5.west |- {0, -2.8}); 
\coordinate (E2_end) at (A2.east |- {0, -2.8});

\draw[<->] (E2_start) -- (E2_end) node[midway, below] {Episode 2};

\end{tikzpicture}
    }
    \caption{Episodes in our \ac{MDP} formulation for LA. Each box represents a \ac{TTI}, and acknowledgments use the same color as the transmission they respond to. Note that episodes 1 and 2 overlap between \ac{TTI} 5 and 12.}
    \label{fig:sequence_1}
\end{figure}
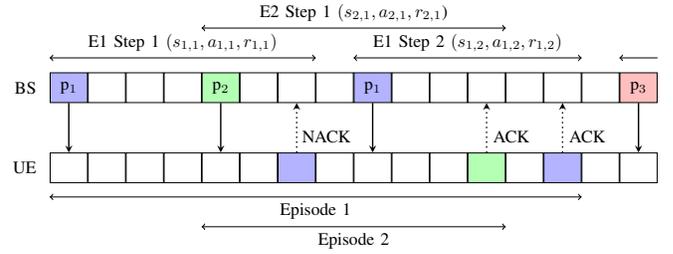
\subsection{\ac{MDP} Formulation}  

We model \ac{LA} as an episodic \ac{MDP} in which an episode (or trajectory) $\tau=(s_0,a_0,r_0,\dots, s_T,a_T,r_T)$  corresponds to the lifespan of a \ac{UE}'s data packet from its first transmission to, alternatively, a successful reception or the packet being dropped after four retransmissions (as illustrated in~\cref{fig:sequence_1}). As such, the episode's length $T$ is bounded to five steps (each corresponding to a transmission of the packet).
\begin{itemize}
    \item The state $s_t \in \mathbb{R}^n$ is a vector of semi-static and dynamic information characterizing the serving cell, the \ac{UE} design, as well as radio measurements, channel state information, \ac{HARQ} feedback, and more\footnote{The exact list of states cannot be disclosed as it is based on Ericsson's internal research work.}.
    \item The action $a_t\in \{0,\dots,27\}$ takes an integer, representing one of 28 \ac{MCS} indices for data transmissions, as defined by the \ac{5G} specifications~\cite{3rd_generation_partnership_project_3gpp_technical_2024}. When dealing with a re-transmission, that is $t\geq 1$,  the simulator maps the agent's action to a reserved \ac{MCS} index in the interval $\{28,\dots,31\}$, as per \ac{5G} specifications.
    \item  The reward signal $r_t=r(s_t,a_t)$ is function of the \ac{SE} of the packet transmission. 
\end{itemize}

Notably, as the \ac{UE}'s \ac{HARQ} process supports multiple active packets at any given time, multiple \ac{UE}'s episodes may overlap in time (as in the example in~\cref{fig:sequence_1}). 

\subsection{\ac{BCQ} and \ac{CQL} for Link Adaptation}
We train both algorithms on static datasets of transitions in the form of $(s_t,a_t,r_{t+1},s_{t+1})$ tuples. The definition of the episode on a packet basis enables agents to learn from experience generated by any \ac{UE} within the network. The formulation detailed above allows adapting \ac{CQL} and \ac{BCQ} to \ac{LA} with minimum effort: in particular, when dealing with a finite set of actions, such as those representing \ac{MCS} indices selectable for \ac{LA}, \ac{BCQ} can overcome the need to explicitly model the distribution of actions in the dataset by simply computing probabilities for each action being selected by the behavioral policy. 
Therefore, inspired by \cite{fujimoto_benchmarking_2019}, we restrict the support of action selection to a subset of actions whose probability exceeds the probability of the action most likely to be selected by the behavioral policy by a minimum threshold. For training a \ac{CQL} policy, instead, we replace the automatic tuning mechanism used in \cite{kumar_conservative_2020} for continuous action spaces by fixing the \textit{tradeoff-factor} to some pre-selected values.

\section{Sequence Modeling for Link Adaptation}\label{sec:method}

The previous \ac{MDP} formulation of \ac{LA}, and thereby the related offline \ac{RL} designs,  present some shortcomings. Firstly, such  formulation optimizes the return for individual packets. Thus, it may not capitalize on the fact that sequences of packets can experience correlated channel conditions that would allow to optimize longer-term metrics (e.g., \ac{UE} throughput). Secondly, since packet arrivals are unknown, and channel conditions are time-varying, the overall stochastic process we aim to control can be more generally framed as a \emph{contextual} \ac{MDP} \cite{hallak2015contextual}. A contextual \ac{MDP} is a specific type of \ac{MDP} that includes an underlying hidden parameter (context), possibly time-varying, affecting the transition dynamics—such as packet arrivals and channel conditions. Managing such a process is generally more complex than controlling a classical \ac{MDP} \cite{kaelbling1998planning}. 

To address these complexities, we reformulate offline \ac{RL} for \ac{LA} as a sequence modeling problem \cite{janner2021offline,emmons_rvs_2021}~and propose a solution that extends and adapts the \ac{DT} design \cite{chen_decision_2021}.

\subsection{RL via Supervised Learning (RvS)}

By framing RL as a sequence modeling problem, we can better capture the time-varying interactions in the underlying process. This approach leverages the strengths of sequence models in handling temporal dependencies.

We consider the \ac{RvS} learning approach for offline RL proposed in \cite{emmons_rvs_2021}. This approach takes as input a dataset ${\cal D}=\{\tau\}$ of trajectories of experience and outputs an \emph{outcome-conditioned} policy $\pi_\theta(a_t\mid s_t,\omega)$ that optimizes
\begin{equation}
    \max_\theta \mathbb{E}_{\tau \sim {\cal D}}\left[\sum_{t=0}^{|\tau|-1} \mathbb{E}_{\omega \sim f(\omega\mid \tau_t)}\left[ \ln \pi_\theta(a_t\mid s_t,\omega)\right]\right],
\end{equation}
where $\tau_t=\{s_t,a_t,r_t,s_{t+1},\dots\}$ denotes the trajectory starting at time $t$, $|\tau|$ denotes the trajectory length, and $f(\omega|\tau)$  denotes the distribution over outcomes $\omega$ that occur in a trajectory. These outcomes are user-chosen, and can be, for example:
\begin{itemize}
    \item \emph{Goal-oriented}: $\omega=\{s_{h}=s\}$, that is, $f(\omega|\tau)$ indicates the likelihood of visiting state $s$ at step $h$.
    \item \emph{Reward-oriented}: $\omega=\sum_{n\geq t}\gamma^{n-t} r_n$, and $f(\omega|\tau)$ indicates the likelihood of achieving a certain total discounted reward starting from $t$. Such conditioning is also referred to as \emph{return-to-go} (RTG) conditioning.
\end{itemize}
At inference time, then, the user can choose to condition the policy according to a desired objective.

\subsection{RvS Learning for LA}
To formulate  \ac{LA}  in terms of the \ac{RvS} learning framework, we introduce a trajectory definition that extends beyond the lifespan of a single packet, departing from the previous \ac{MDP} formulation.  After that, we introduce different ways to condition the policy $\pi_\theta$.

\subsubsection{Trajectory definition} \label{trajectory definition} We consider fixed-horizon trajectories $\tau=\{s_0,a_0,r_0,\dots, s_H,a_H,r_H\}$, which include  transmissions of packets belonging to a specific \ac{UE}.

To construct $\tau$ we consider two alternative approaches: 

\begin{enumerate}[label=\alph*)]
    \item \emph{Recent transmissions}:  $\tau$ consists of the most recent $K$ transmissions of the \ac{UE} considered (hence $H=K$).
    \item \emph{Consecutive packets}: $\tau$ includes only up to  $n_p$ of a \ac{UE}'s most recent consecutive packets. Since a packet can be retransmitted at most four times upon its initial transmission, we have that $H\in\{n_p,\dots, 5n_p\}$.
\end{enumerate}
At training time, we sample trajectories of data from ${\cal D}$  according to one of these two approaches, and, if needed, extra positions in $\tau$ are filled using padding tokens.



\subsubsection{Outcome conditioning}\label{subsec:conditioning} We propose the following three ways to condition the policy $\pi_\theta$ for \ac{LA}:
\begin{enumerate}[label=\alph*)]
    \item {\bf VANILLA}: A simple conditioning consisting in fixing a target \ac{RTG} value $\omega$ for each packet $p$. At training time, the \ac{RTG} at the $k$-th step of a packet is computed as  $\omega=\sum_{t=k}^{T(p)} \gamma^{t-k} r_k(p)$, where $r_k(p)$ indicates the reward associated to the $k$-th transmission for packet $p$, $\gamma \in(0,1]$, and  $T(p)$ is a random variable indicating the termination step for the transmission of packet $p$. At inference time, the \ac{RTG} is fixed to a pre-defined value at the first transmission and adjusted in case of retransmissions (we explore this in~\cref{sec:experiments}).

    \item {\bf \Ac{DAVG}}: The VANILLA approach has the same shortcomings of the MDP formulation proposed in \cref{sec:problem}. An alternative is to condition according to a \ac{DAVG} of  future rewards. Letting $n$ denote the $n$-th step in a sequence of continuous transmissions from a UE, at training time we condition according to:
    \[
           \omega=(1-\gamma)\sum_{t=n}^\infty \gamma^{t-n}r_t,
    \]
    with $\gamma \in(0,1)$.   At inference time, as for the VANILLA case, we condition according to some pre-defined value.
    \item {\bf \Ac{CCTR}}:  As the definition of the reward signal (cf.~\cref{sec:problem}) directly relates to the channel's spectral efficiency, it may be possible that some values of spectral efficiency are unattainable depending on the channel's condition. 
    To address this issue, we propose a novel conditioning that dynamically adapts the policy $\pi_\theta$ based on the observed channel quality. For the first packet transmission, we decide to use as \ac{RTG} target value the nominal spectral efficiency associated with the latest \ac{CQI} value reported by the \ac{UE} (according to Table 5.2.2.1\text{-}3 of the \ac{3GPP} specification~\cite{3rd_generation_partnership_project_3gpp_technical_2024}).
    
\end{enumerate}

\subsection{Decision Transformers Design for LA}

We next describe how we adapted the \ac{DT} architecture for the link adaptation problem. We introduce two changes: one to the attention layer, and one to the positional embedding.
\begin{figure}[t]
    \centering
    \resizebox{\linewidth}{!}{%
    \begin{tikzpicture}

\tikzstyle{block} = [draw, minimum width=0.75cm, minimum height=0.6cm, align=center]
\tikzstyle{arrow} = [thick,->,>=stealth]
\definecolor{darkgreen}{RGB}{0, 150, 0} 

\node[block, fill=blue!30] (P1) {p$_1$};
\node[block, right=0cm of P1] (E1B2) {};
\node[block, right=0cm of E1B2] (E1B3) {};
\node[block, right=0cm of E1B3] (E1B4) {};
\node[block, right=0cm of E1B4, fill=green!30] (P2) {p$_2$};
\node[block, right=0cm of P2] (E1B5) {};
\node[block, right=0cm of E1B5] (E1B6) {};
\node[block, right=0cm of E1B6] (E1B7) {};
\node[block, right=0cm of E1B7, fill=orange!30] (P3) {p$_3$};
\node[block, right=0cm of P3] (E1B8) {};
\node[block, right=0cm of E1B8] (E1B9) {};
\node[block, right=0cm of E1B9] (E1B10) {};
\node[block, right=0cm of E1B10] (E1B11) {};
\node[block, right=0cm of E1B11] (E1B12) {};
\node[block, right=0cm of E1B12] (E1B13) {};
\node[block, right=0cm of E1B13, fill=pink] (P4) {p$_4$};

\node[block, below=1cm of P1] (E2B1) {};
\node[block, right=0cm of E2B1] (E2B2) {};
\node[block, right=0cm of E2B2] (E2B3) {};
\node[block, right=0cm of E2B3] (E2B4) {};
\node[block, right=0cm of E2B4] (E2B5) {};
\node[block, right=0cm of E2B5] (E2B6) {};
\node[block, fill=green!30, right=0cm of E2B6] (A1) {};
\node[block, right=0cm of A1] (E2B8) {};
\node[block, right=0cm of E2B8] (E2B9) {};
\node[block, right=0cm of E2B9] (E2B10) {};
\node[block, right=0cm of E2B10] (E2B11) {};
\node[block, fill=blue!30, right=0cm of E2B11] (A2) {};
\node[block, right=0cm of A2] (E2B13) {};
\node[block, fill=orange!30, right=0cm of E2B13] (A3) {};
\node[block, right=0cm of A3] (E2B15) {};
\node[block, right=0cm of E2B15] (E2B16) {};

\node[left=0.15cm of P1] (BS) {BS};
\node[left=0.15cm of E2B1] (UE) {UE};

\node[above=0.08cm of A1, shift={(0.6cm, 0)}] (NACK) {NACK};
\node[above=0.08cm of A2, shift={(0.5cm, 0)}] (ACK) {ACK};
\node[above=0.08cm of A3, shift={(0.5cm, 0)}] (ACK2) {ACK};

\draw[arrow, dotted] (A1.north) -- (E1B6.south);
\draw[arrow, dotted] (A2.north) -- (E1B10.south);
\draw[arrow, dotted] (A3.north) -- (E1B12.south);

\draw[arrow] (P1.south) -- (E2B1.north);
\draw[arrow] (P2.south) -- (E2B5.north);
\draw[arrow] (P3.south) -- (E2B9.north);
\draw[arrow] (P4.south) -- (E2B16.north);


\draw[dotted, rounded corners=5pt, fill=yellow, fill opacity=0.1] (-1,-2.5) rectangle (12,-8.5);

\coordinate (MASK) at (5.5, -3);  

\node[below=-0.2cm of MASK, shift={(0cm, 0)}, font=\Large] (c) {Attention masking};

\tikzstyle{block} = [draw, minimum width=1cm, minimum height=1cm, align=center]

\node[block, fill=white] (x) at (1, -4.5) (1) {1};
\node[block, shift={(0cm, -1cm)}, fill=white] at (1, -4.5) (2) {1};
\node[block, shift={(0cm, -2cm)},fill=white] at (1, -4.5) (3) {1};
\node[block, shift={(0cm, -3cm)}, fill=white] at (1, -4.5) (4) {1};
\node[block, shift={(1cm, 0cm)}, fill=darkgreen!50] at (1, -4.5) (5) {0};
\node[block, shift={(1cm, -1cm)}, fill=white] at (1, -4.5) {1};
\node[block, shift={(1cm, -2cm)}, fill=white] at (1, -4.5) {1};
\node[block, shift={(1cm, -3cm)}, fill=white] at (1, -4.5) {1};
\node[block, shift={(2cm, 0cm)}, fill=darkgreen!50] at (1, -4.5) (6) {0};
\node[block, shift={(2cm, -1cm)}, fill=darkgreen!50] at (1, -4.5)  {0};
\node[block, shift={(2cm, -2cm)}, fill=white] at (1, -4.5) {1};
\node[block, shift={(2cm, -3cm)}, fill=white] at (1, -4.5) {1};
\node[block, shift={(3cm, -0cm)}, fill=darkgreen!50] at (1, -4.5) (7) {0};
\node[block, shift={(3cm, -1cm)}, fill=darkgreen!50] at (1, -4.5) (dimarco) {0};
\node[block, shift={(3cm, -2cm)}, fill=darkgreen!50] at (1, -4.5) (pavard) {0};
\node[block, shift={(3cm, -3cm)}, fill=white] at (1, -4.5) {1};

\node[left=0.08cm of 1] (p1) {p$_1$};
\node[left=0.08cm of 2] (p2) {p$_2$};
\node[left=0.08cm of 3] (p3) {p$_3$};
\node[left=0.08cm of 4] (p4) {p$_4$};

\node[above=0.08cm of 1] (p1) {p$_1$};
\node[above=0.08cm of 5] (p2) {p$_2$};
\node[above=0.08cm of 6] (p3) {p$_3$};
\node[above=0.08cm of 7] (p4) {p$_4$};

\node[block, fill=white] (x) at (7, -4.5) (1) {1};
\node[block, shift={(0cm, -1cm)}, fill=red!50] at (7, -4.5) (2) {0};
\node[block, shift={(0cm, -2cm)}, fill=red!50] at (7, -4.5) (3) {0};
\node[block, shift={(0cm, -3cm)}, fill=white] at (7, -4.5) (4) {1};
\node[block, shift={(1cm, 0cm)}, fill=darkgreen!50] at (7, -4.5) (5) {0};
\node[block, shift={(1cm, -1cm)}, fill=white] at (7, -4.5) {1};
\node[block, shift={(1cm, -2cm)}, fill=white] at (7, -4.5) {1};
\node[block, shift={(1cm, -3cm)}, fill=white] at (7, -4.5) {1};
\node[block, shift={(2cm, 0cm)}, fill=darkgreen!50] at (7, -4.5) (6) {0};
\node[block, shift={(2cm, -1cm)}, fill=darkgreen!50] at (7, -4.5) {0};
\node[block, shift={(2cm, -2cm)}, fill=white] at (7, -4.5) {1};
\node[block, shift={(2cm, -3cm)}, fill=white] at (7, -4.5) {1};
\node[block, shift={(3cm, -0cm)}, fill=darkgreen!50] at (7, -4.5) (7) {0};
\node[block, shift={(3cm, -1cm)}, fill=darkgreen!50] at (7, -4.5) {0};
\node[block, shift={(3cm, -2cm)}, fill=darkgreen!50] at (7, -4.5) {0};
\node[block, shift={(3cm, -3cm)}, fill=white] at (7, -4.5) {1};

\node[left=0.08cm of 1] (p1) {p$_1$};
\node[left=0.08cm of 2] (p2) {p$_2$};
\node[left=0.08cm of 3] (p3) {p$_3$};
\node[left=0.08cm of 4] (p4) {p$_4$};

\node[above=0.08cm of 1] (p1) {p$_1$};
\node[above=0.08cm of 5] (p2) {p$_2$};
\node[above=0.08cm of 6] (p3) {p$_3$};
\node[above=0.08cm of 7] (p4) {p$_4$};

\draw[->, line width=0.4mm, black, shorten >=8pt, shorten <=8pt] (dimarco.south east) -- (3.north west);

\end{tikzpicture}
    }
    \caption{\textbf{At training time}, the proposed attention mask (right matrix) prevents the tokens associated to $p_2$ and $p_4$ from attending $p_1$, as its reward is not available at the time an MCS index is requested for $p_2$ and $p_4$. The left matrix shows the original attention mask definition.}
    \label{fig: attn mask}
\end{figure}
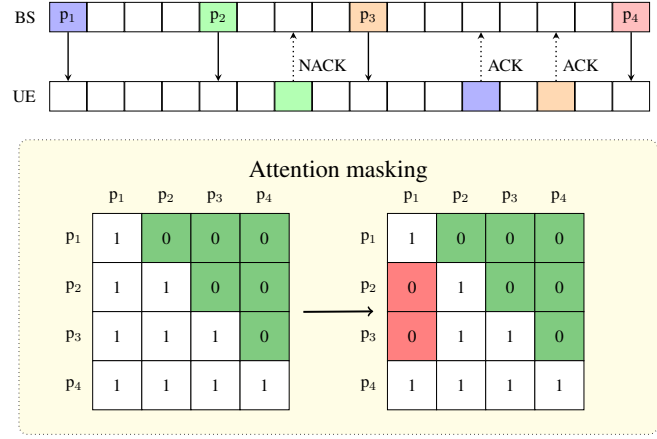


\subsubsection{Attention masking for delayed feedback}

A significant challenge in training a \ac{DT} architecture for \ac{LA} is preventing the leakage of future data during the training process. Specifically, the \ac{DT}'s inference outputs must rely solely on information accessible as in an actual deployment, avoiding any use of future data that would not be accessible in real-time. As historical rewards are only available upon a base station receiving a \ac{HARQ} feedback (positive or negative) for previous packet transmissions, we propose an attention-masking approach enforcing causality while coping with such delayed feedback. To this end, we model a trajectory~$\tau$~as: \[\tau=(\underbrace{s_0}_{X_{0}},\underbrace{a_0}_{X_1},\underbrace{r_0}_{X_2},\dots, \underbrace{s_K}_{X_{3K}}, \underbrace{a_K}_{X_{3K+1}}, \underbrace{r_K}_{X_{3K+2}}),\]
where $(X_0, X_1,\dots, X_{3K+2})$ indicates the associated sequence of input tokens, and we define $t_a(p)$ and $t_r(p)$ as the time at which, respectively,  the action for packet $p$ is scheduled and the associated reward signal is available. The proposed attention mask between any two tokens $X_i,X_j $ is:
\[
\hbox{\texttt{attn\_mask}}(X_i,X_j) = \mathbf{1}\{ t_r(p(X_i)) \leq t_a(p(X_j))\},
\]
where $p(X)$ is a mapping between a token $X$ and the associated packet $p$.
As illustrated in \cref{fig: attn mask}, the mask assures that if a reward for a packet $p(X_i)$ is not received when an action for a packet $p(X_j)$ is taken, then tokens of packet $p(X_j)$ are not affected by the tokens of packet $p(X_i)$.

\begin{figure*}[ht]
    \centering
    \begin{minipage}{0.30\textwidth}
        \centering
        \includegraphics[trim=380 0 0 5, clip, width=\textwidth]{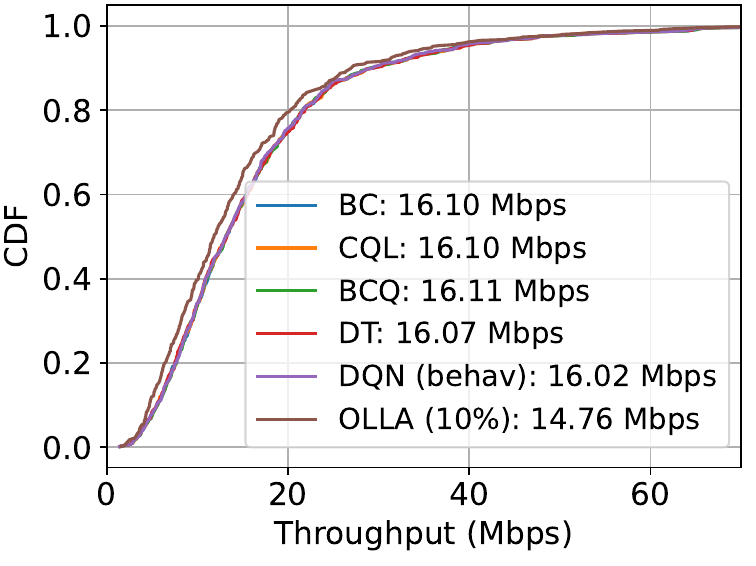}
       
    \end{minipage}%
    \hfill
    \begin{minipage}{0.30\textwidth}
        \centering
        \includegraphics[trim=380 0 0 5, clip, width=\textwidth]{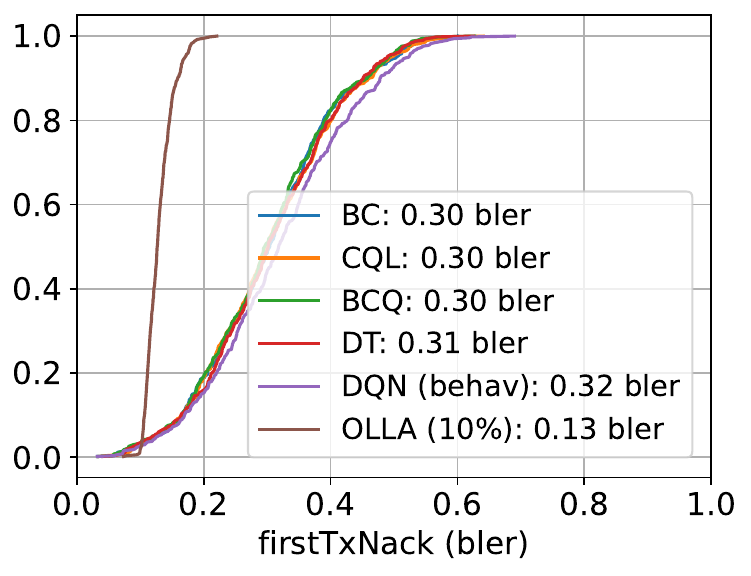}
       
    \end{minipage}%
    \hfill
    \begin{minipage}{0.30\textwidth}
        \centering
        \includegraphics[trim=380 0 0 5, clip, width=\textwidth]{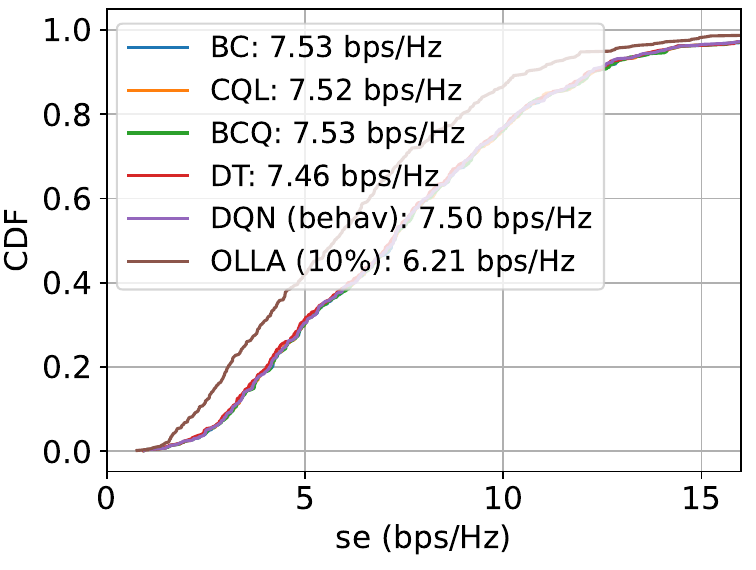}
      
    \end{minipage}
     \caption{Cumulative distribution function (CDF) of user throughput, \ac{BLER} and \ac{SE} for various offline \ac{RL} models trained on $\mathcal{D}_{\rm opt}$. All offline \ac{RL} algorithms achieve similar performance across all metrics, comparable with the online \ac{DQN} policy, and outperform the \ac{OLLA} baseline.}
     \label{cdf}
\end{figure*}

\subsubsection{Temporal embeddings}
\ac{LA} is a real-time process with states, actions, and rewards observed at irregular intervals, whereas the \ac{DT} architecture assumes equidistant sampling. To handle irregular observations, we propose three alternative definitions of positional encoding attempting to  capture temporal correlations between successive packet transmissions of a \ac{UE}:
%
\begin{enumerate}
    \item \textbf{\Ac{BE}}: as a baseline, we implement a minor modification to the original encoder from \cite{chen_decision_2021}, where embeddings are learned based on each token's index within the sequence (i.e., $i \in [0, K-1]$).
    \item \textbf{\Ac{LT} encoding}: 
    embeddings are learned using a linear layer with $(\Delta {t}_i)_{i=0}^K$ as input, where $\Delta t_i= t_a(p(X_i)) - t_a(p(X_0))$. Hence, $\Delta {t}_i$ is a measure of the relative time  between the $i$-th token and the first one.
    Note that tokens within the same packet have identical relative timestamps, allowing the \ac{DT} to associate the reward to the corresponding state and action.

    \item \textbf{\Ac{CT} encoding}: inspired by \cite{zuo2020transformer, zhang2023continuous},  we compute the $k$-th element of the positional embedding vector for token $X_i$ as:
\begin{equation}
    \textup{PE}(\Delta t_i, k) = \begin{cases}
    \sin\left(\displaystyle \frac{\Delta t_i / 100}{{C}^{\frac k {d_\text{model}}}}\right) & \text{ for even } k, \\[4ex]
    \cos\left(\displaystyle \frac {\Delta t_i / 100} {C^{\frac {k-1} {d_\text{model}}}}\right) & \text{ for odd } k,
    \end{cases}
    \end{equation}
    where $d_{\text{model}}$ is a user-chosen constant (see also  \cite{zhang2023continuous}).
\end{enumerate}

\section{Numerical Results}\label{sec:experiments}

This section compares the various offline \ac{RL} designs for \ac{LA} using an industry-grade simulator. Specifically, we investigate: (1) whether offline \ac{RL} methods can match the performance of the behavioral policy and \ac{OLLA}; (2) how the behavioral policy affects the performance of these methods; and (3), the impact of different design aspects of the \ac{DT} on \ac{LA} performance.

\subsection{Environment and Data Collection}
\label{subsec:collection}

\subsubsection{Environment} We rely on an Ericsson's proprietary \ac{RAN} simulator platform compliant with the \ac{3GPP} \ac{5G} technical specification. Specifically, we consider a \ac{TDD} 5G system operating at a 3.5GHz carrier frequency, with PHY layer numerology $\mu=0$ (cf. \ac{3GPP} TS 38.200 family) and \ac{SU-MIMO} transmission. Each simulation consists of a tri-sectorial site, with each sector served by a \ac{mMIMO} 8x4x2 antenna array, and 10 UEs with full buffer traffic randomly dropped in the site coverage area. Each simulation (i.e., seed) simulates 5 seconds (i.e., $\approx \ $5000 \acp{TTI}).

\subsubsection{Data collection}
To investigate the potential of offline \ac{RL} to achieve optimal performance (e.g., comparable with a policy learned online) when trained with proper data,  we collect training data using two \ac{DQN} policies pre-trained online: A \ac{DQN} policy $\pi_{\rm opt}$ trained until convergence; and a sub-optimal \ac{DQN} policy $\pi_{\rm s-opt}$ trained with half the data of $\pi_{\rm opt}$. The latter allows us to analyze whether offline \ac{RL} algorithms can infer optimal behavior from sub-optimal training data. For each policy, we collected data across $20$ seeds, resulting into two datasets, respectively, $\mathcal{D}_{\rm opt}$ and $\mathcal{D}_{\rm s-opt}$, of $\approx 400k$ transitions. 

\begin{table*}[t!]
\centering
\begin{minipage}{0.48\linewidth}  
    \centering
    \caption{Comparison of the offline RL methods trained on $\mathcal{D}_{\rm opt}$. \\ OLLA (x\%) refers to OLLA with a fixed BLER target of x\%.}
    \setlength{\tabcolsep}{8pt} 
    \begin{tabular}{lccc}
    \hline
    \toprule\toprule
        \ac{LA} algorithm & Throughput & \ac{SE} & \ac{BLER} \\
    \toprule
        OLLA ($10\%$) & 14.76 Mbps   & 6.21 bps/Hz   & 0.13     \\
        OLLA ($30\%$) & 15.51 Mbps   & 6.87 bps/Hz   & 0.30     \\
        OLLA ($50\%$) & 15.16 Mbps   & 6.92 bps/Hz   & 0.49     \\
        OLLA ($70\%$) & 14.77 Mbps   & 6.81 bps/Hz   & 0.66     \\
        OLLA ($90\%$) & 14.71 Mbps   & 6.84 bps/Hz   & 0.83     \\
        DQN$_{\rm opt}$   & 16.02 Mbps   & 7.50 bps/Hz   & 0.32     \\
        BC            & 16.10 Mbps   & 7.53 bps/Hz   & 0.30     \\
        BCQ           & 16.11 Mbps   & 7.53 bps/Hz   & 0.30     \\
        CQL           & 16.10 Mbps   & 7.52 bps/Hz   & 0.30     \\
        DT            & 16.07 Mbps   & 7.46 bps/Hz   & 0.31     \\
    \bottomrule\bottomrule
    \hline
    \end{tabular}
    \vspace{0.25cm} 
    \label{table: opt table}
\end{minipage}
\hspace{0.02\linewidth}  
\begin{minipage}{0.48\linewidth}  
    \centering
    \caption{Comparison of the offline RL methods trained on $\mathcal{D}_{\rm s-opt}$. \\ OLLA (x\%) refers to OLLA with a fixed BLER target of x\%.}
    \setlength{\tabcolsep}{8pt} 
    \begin{tabular}{lccc}
    \hline
    \toprule\toprule
        \ac{LA} algorithm & Throughput & \ac{SE} & \ac{BLER} \\
    \toprule
        OLLA ($10\%$) & 14.76 Mbps   & 6.21 bps/Hz   & 0.13     \\
        OLLA ($30\%$) & 15.51 Mbps   & 6.87 bps/Hz   & 0.30     \\
        OLLA ($50\%$) & 15.16 Mbps   & 6.92 bps/Hz   & 0.49     \\
        OLLA ($70\%$) & 14.77 Mbps   & 6.81 bps/Hz   & 0.66     \\
        OLLA ($90\%$) & 14.71 Mbps   & 6.84 bps/Hz   & 0.83     \\
        DQN$_{\rm s-opt}$ & 15.89 Mbps  & 7.45 bps/Hz   & 0.33 \\
        BC            & 16.00 Mbps   & 7.50 bps/Hz   & 0.31     \\
        BCQ           & 16.00 Mbps   & 7.60 bps/Hz   & 0.34     \\
        CQL           & 16.05 Mbps   & 7.59 bps/Hz   & 0.33     \\
        DT            & 15.67 Mbps   & 7.25 bps/Hz   & 0.36     \\
    \bottomrule\bottomrule
    \hline
    \end{tabular}
    \vspace{0.25cm} 
    \label{table: s-opt table}
\end{minipage}
\end{table*}

\subsection{General Results}

We compare the performance of our \ac{LA} designs based on value-based offline \ac{RL} (i.e., \ac{BCQ} \cite{fujimoto2019off} and \ac{CQL} \cite{kumar_conservative_2020}), and sequence modeling with \ac{DT}. We evaluate their performance in terms of average \ac{UE} throughput, \ac{SE} and experienced \ac{BLER}. We consider three baselines: the \ac{OLLA} approach introduced in~\cref{sec:problem formulation}, representing the state-of-the-art in real-world \acp{RAN}; a simple \ac{BC} approach, which learns to reproduce training behavior;  and the behavioral \ac{DQN} policy. We tested the algorithm using $50$ randomized scenarios; for additional design choices and extended descriptions, we refer the reader to \ifdefined\addappendix the appendix \else the technical report~\cite{Peri:24}. \fi

\cref{cdf} presents the \acp{CDF} of \ac{UE} throughput, \ac{SE} and experienced \ac{BLER} for the best configuration of each offline \ac{RL} \ac{LA} algorithm resulting from experimentation with different design choices and hyperparameters, with models trained on $\mathcal{D}_{\rm opt}$. For the \ac{DT} design, this implies sampling sequences including the most recent $H=32$ transitions and following the \ac{CCTR} conditioning approach. Furthermore, \cref{table: opt table,table: s-opt table} summarize, respectively, the average results for all models trained using $\mathcal{D}_{\rm opt}$ and $\mathcal{D}_{\rm s-opt}$.

When training on $\mathcal{D}_{\rm opt}$, \cref{table: opt table} shows that all offline \ac{RL} methods achieve similar results across all the metrics, reaching the performance of the data collection policy $\pi_{\rm opt}$ and outperforming the industry baseline \ac{OLLA} by $\approx 8.8\%$ in average \ac{UE} throughput and $\approx 20.7\%$ in average \ac{UE} \ac{SE}.

With sub-optimal training data  $\mathcal{D}_{\rm s-opt}$, \cref{table: s-opt table} shows that \ac{CQL} and \ac{BCQ} reach approximately the same performance observed when trained on $\mathcal{D}_\text{opt}$, on pair with the behavioral policy $\mathrm{DQN}_{\rm s-opt}$ and \ac{BC}. However, in this case \ac{DT} shows a slight degradation compared to $\mathrm{DQN}_{\rm s-opt}$, but consistently outperforms the OLLA baseline.

\subsection{Considerations on the \ac{DT} Design}
Hereafter, we analyze in more details the effect of different design aspects of the \ac{DT}. \cref{table:pe and conditioning - throughput} summarizes the best results obtained for different combinations of the positional encoder and conditioning methods presented in \cref{sec:method}. For this analysis, we considered $\mathcal{D}_\text{opt}$ as training set and trajectories constructed with the \emph{recent transmissions} approach introduced in \cref{trajectory definition}, with $H=32$. For the VANILLA and DAVG conditioning, we conditioned on the 90\% quantile of the distribution of RTGs in the training dataset (filtered at the first retransmission in the case of VANILLA conditioning). The average future rewards are computed over a fixed window size of $35$ steps, with a discount factor $\gamma = 0.8$. 

For positional encoding, our results indicate that including temporal information about packet distribution is beneficial. Interestingly, the \ac{LT} approach, which learns embeddings directly from the actual timestamps of packet transmissions, achieves better performance than \ac{CT}, which computes encodings with trigonometric functions, and also the baseline.

Furthermore, considering \ac{RTG} conditioning, our results in \cref{table:pe and conditioning - throughput} indicate that \ac{CCTR} conditioning achieves better throughput than both VANILLA and DAVG conditioning across all types of positional encoding. We observed similar results in terms of \ac{BLER} and \ac{SE}, which are not included here for brevity. This demonstrates that, for the \ac{LA} problem, conditioning the model inference based on the channel state is important. 

\begin{table}[]
\centering
\caption{Comparison of different types of position encoding and RTG conditioning. Best in \textbf{bold}, second best in red.
}
\renewcommand{\arraystretch}{1.25}
\begin{tabular}{lccc}
\toprule\toprule
Pos. Emb.     & VANILLA       & CCTR          & DAVG           \\ \hline
BE          & 15.75 Mbps & 15.83 Mbps & 15.76 Mbps \\
CT              & 15.35 Mbps & 15.86 Mbps & 15.73 Mbps \\
LT & \textcolor{red}{16.01 Mbps} & \textbf{16.07 Mbps} & 15.66 Mbps \\
\bottomrule\bottomrule
\end{tabular}
\label{table:pe and conditioning - throughput}
\end{table}

\subsection{Outcome Conditioning}
\label{RTG conditioning}

We further analyze the effect of output conditioning for \ac{LA} with respect to (1) the type of conditioning (using the VANILLA and \ac{CCTR} approaches, cf. \cref{subsec:conditioning}); and (2) the sequence length (following the \emph{consecutive packets} approach).
\Cref{table: more-packets} summarizes results that consider input sequences that comprise transmissions related to $n_p\in[1,\; 16]$ packets and, for VANILLA conditioning, three \ac{RTG} target values $\omega$ chosen as the $\{20,60,100\}$-quantiles of the training \ac{RTG} distribution.


For wireless applications, like \ac{LA}, conditioning the action generation to \ac{RTG} target values that reflect the radio conditions is instrumental in achieving good performance. This is demonstrated by the \ac{CCTR} approach, which adapts the \ac{RTG} target value for each packet transmission using the nominal spectral efficiency corresponding to the latest \ac{CQI} reported by the \ac{UE}. On the other hand, conditioning the action generation using fixed \ac{RTG} target values that do not reflect the radio conditions, as in the VANILLA approach, can lead to substantial performance degradation. For instance, choosing a too high (fixed) \ac{RTG} target value causes the \ac{DT} to take more aggressive \ac{LA} actions, i.e., higher \ac{MCS} index, which leads to an increased frequency of retransmissions during testing, reflected by lower throughput and \ac{SE} and a higher \ac{BLER}.

Although the transformer architecture excels in learning from long sequences of information, for wireless applications that act directly on packet transmissions, like \ac{LA}, long sequences appear not to be desirable. In fact, when consecutive tokens in the sequence are associated with data packets transmitted far apart in time (e.g., more than the channel coherence time), learning auto-regressively from too-aged historical tokens can induce variance and cause the \ac{DT} to revert to behavioral cloning. For instance, \cref{table: more-packets} shows that, with \ac{CCTR} conditioning, the \ac{DT} benefits from including more packets in the sequence up until $n_p=8$, longer sequences do not yield better performance and \ac{BLER} values converge to the one achieved by the behavioral \ac{DQN}. We refer to the \ifdefined\addappendix appendix \else technical report~\cite{Peri:24} \fi for an extended analysis.


\begin{table}[t!]
    \centering
    \setlength{\tabcolsep}{3pt}
    \caption{Performance obtained by conditioning on increasing, fixed, RTG values at inference for models including variable numbers of packets inside their sequence (VANILLA and \ac{CCTR}).}
    \begin{tabular}{llllll}
    \toprule\toprule
    $n_p$ & Metrics      & CCTR     & $\omega=1.71$  & $\omega=3.86$ & $\omega=7.57$ \\ \hline
         & Th.put    & 15.73 Mbps & 15.30 Mbps & 13.80 Mbps & 13.94 Mbps \\
    1    & BLER          & 0.37       & 0.33       & 0.42       & 0.83       \\
         & SE            & 7.45 bps/Hz & 7.14 bps/Hz & 6.92 bps/Hz & 7.19 bps/Hz \vspace{1.5mm} \\
         & Th.put    & 15.72 Mbps & 15.08 Mbps & 13.75 Mbps & 13.98 Mbps \\
    2    & BLER          & 0.38       & 0.31       & 0.41       & 0.82       \\
         & SE            & 7.45 bps/Hz & 7.08 bps/Hz & 6.87 bps/Hz & 7.25 bps/Hz \vspace{1.5mm} \\
         & Th.put    & 16.02 Mbps & 15.93 Mbps & 15.45 Mbps & 14.06 Mbps \\
    4    & BLER          & 0.31       & 0.28       & 0.36       & 0.74       \\
         & SE            & 7.43 bps/Hz & 7.32 bps/Hz & 7.45 bps/Hz & 7.46 bps/Hz \vspace{1.5mm} \\
         & Th.put    & 16.04 Mbps & 16.01 Mbps & 15.96 Mbps & 14.49 Mbps \\
    8    & BLER          & 0.28       & 0.27       & 0.30       & 0.62       \\
         & SE            & 7.40 bps/Hz & 7.34 bps/Hz & 7.42 bps/Hz & 7.56 bps/Hz \vspace{1.5mm} \\
         & Th.put    & 16.02 Mbps & 15.93 Mbps & 15.45 Mbps & 14.06 Mbps \\
    12   & BLER          & 0.31       & 0.30       & 0.31       & 0.46       \\
         & SE            & 7.45 bps/Hz & 7.42 bps/Hz & 7.46 bps/Hz & 7.58 bps/Hz \vspace{1.5mm} \\
         & Th.put    & 16.02 Mbps & 15.93 Mbps & 15.45 Mbps & 14.06 Mbps \\
    16   & BLER          & 0.30       & 0.29       & 0.31       & 0.40       \\
         & SE            & 7.44 bps/Hz & 7.42 bps/Hz & 7.44 bps/Hz & 7.48 bps/Hz \\
    \bottomrule\bottomrule
    \end{tabular}
    \label{table: more-packets}
\end{table}

\section{Final Remarks}
\label{sec:conclusion}

In this paper, we designed three offline \ac{RL} algorithms, aiming towards replacing the legacy \ac{OLLA} scheme with a scheme that learns from data without impacting the network performance negatively. Specifically, we investigated the performance of \ac{BCQ}, \ac{CQL}, and a \ac{DT}-based design, with a trained \ac{DQN} model serving as the behavioral policy for data collection. 
To gain insight into the effect of data collection, we used two behavioral policies, both trained with \ac{DQN}: (1) a policy $\pi_{\rm opt}$ and (2) a policy $\pi_{\rm s-opt}$, which was trained on half of the data used for training  $\pi_{\rm opt}$.
For the \ac{DT}-based design, we have modified (1) the associated attention layer by suitable masking and (2) the positional encoding definition in order to capture the temporal correlation between subsequent packet transmissions.

The results indicate that the three offline \ac{RL} schemes considered in this work yield similar \ac{LA} performance in terms of the achieved throughput, \ac{BLER} and \ac{SE}, which are also similar to those achieved by the online \ac{DQN} policy and outperform the industry baseline \ac{OLLA} 
in terms of average \ac{UE} throughput (by $\approx 8.8\%$) and average \ac{SE} (by $\approx 20.7\%$).
Although there is no significant gap in the performance of the three offline \ac{RL} methods, our evaluation focuses on a specific network deployment scenario. In this context, all algorithms can benefit from relatively small models, particularly \ac{BCQ} and \ac{CQL}, resulting in lower computational complexity compared to \ac{DT}.
However, achieving model generalization over the \ac{RAN} environment, and thereby being able to deploy the same AI model across cells in large geographical areas, would require training models on substantially larger data sets collected across the network deployment. As such, the ability of the transformer architecture to effectively learn and extrapolate intricate patterns from large datasets may lead to better generalization and performance. Further researching these aspects constitutes an important future work.

Finally, although a \ac{DQN}-based data collection is helpful for investigating whether offline  \ac{RL}  can achieve similar  \ac{LA}  performance as that achieved by online \ac{RL}, finding a non-intrusive data collection policy that properly explores the state-action space without affecting the system performance in live \acp{RAN} remains an open problem, which we are considering in our future research.

\printbibliography

@article{janner2021offline,
  title={Offline reinforcement learning as one big sequence modeling problem},
  author={Janner, Michael and Li, Qiyang and Levine, Sergey},
  journal={Advances in neural information processing systems},
  volume={34},
  pages={1273--1286},
  year={2021},
}

@article{duran_self-optimization_2015,
	title = {Self-optimization algorithm for outer loop link adaptation in {LTE}},
	volume = {19},
	issn = {1089-7798},
	number = {11},
	journal = {IEEE Communications letters},
	author = {Duran, A and Toril, Matías and Ruiz, Fernando and Mendo, Adriano},
	year = {2015},
	note = {Publisher: IEEE},
	pages = {2005--2008},
}

@article{hallak2015contextual,
  title={Contextual {Markov} decision processes},
  author={Hallak, Assaf and Di Castro, Dotan and Mannor, Shie},
  journal={arXiv preprint arXiv:1502.02259},
  year={2015}
}

@article{kaelbling1998planning,
  title={Planning and acting in partially observable stochastic domains},
  author={Kaelbling, Leslie Pack and Littman, Michael L and Cassandra, Anthony R},
  journal={Artificial intelligence},
  volume={101},
  number={1-2},
  pages={99--134},
  year={1998},
  publisher={Elsevier}
}

@book{puterman2014markov,
  title={Markov decision processes: discrete stochastic dynamic programming},
  author={Puterman, Martin L},
  year={2014},
  publisher={John Wiley \& Sons}
}

@inproceedings{mollerstedt2024model,
  title={Model based residual policy learning with applications to antenna control},
  author={M{\"o}llerstedt, Viktor Eriksson and Russo, Alessio and Bouton, Maxime},
  booktitle={2024 IEEE International Conference on Machine Learning for Communication and Networking (ICMLCN)},
  pages={405--411},
  year={2024},
  organization={IEEE}
}

@article{russo2024model,
  title={Model-free active exploration in reinforcement learning},
  author={Russo, Alessio and Proutiere, Alexandre},
  journal={Advances in Neural Information Processing Systems},
  volume={36},
  year={2024}
}

@inproceedings{gronland2024constrained,
  title={Constrained deep reinforcement learning for fronthaul compression optimization},
  author={Gr{\"o}nland, Axel and Russo, Alessio and Jedra, Yassir and Klaiqi, Bleron and Gelabert, Xavier},
  booktitle={2024 IEEE International Conference on Machine Learning for Communication and Networking (ICMLCN)},
  pages={498--504},
  year={2024},
  organization={IEEE}
}

@article{blanquez-casado_eolla_2016,
	title = {{eOLLA}: an enhanced outer loop link adaptation for cellular networks},
	volume = {2016},
	journal = {EURASIP Journal on Wireless Communications and Networking},
	author = {Blanquez-Casado, Francisco and Gomez, Gerardo and Aguayo-Torres, Maria del Carmen and Entrambasaguas, Jose Tomas},
	year = {2016},
	note = {Publisher: Springer},
	pages = {1--16},
}

@article{calabrese_learning_2018,
	title = {Learning radio resource management in {RANs}: {Framework}, opportunities, and challenges},
	volume = {56},
	issn = {0163-6804},
	number = {9},
	journal = {IEEE Communications Magazine},
	author = {Calabrese, Francesco Davide and Wang, Li and Ghadimi, Euhanna and Peters, Gunnar and Hanzo, Lajos and Soldati, Pablo},
	year = {2018},
	note = {Publisher: IEEE},
	pages = {138--145},
}

@article{generalization_2024,
	title = {Design principles for model generalization and scalable {AI} integration in radio access networks},
	issn = {0163-6804},
	number = {9},
	journal = {IEEE Communications Magazine},
	author = {Pablo Soldati and Euhanna Ghadimi and Burak Demirel and Yu Wang and Raimundas Gaigalas and Mathias Sintorn},
	year = {2024},
	note = {Publisher: IEEE},
	pages = {1-8},
}

@article{mnih_playing_2013,
	title = {Playing {A}tari with deep reinforcement learning},
	journal = {arXiv preprint arXiv:1312.5602},
	author = {Mnih, Volodymyr and Kavukcuoglu, Koray and Silver, David and Graves, Alex and Antonoglou, Ioannis and Wierstra, Daan and Riedmiller, Martin},
	year = {2013},
}

@article{levine_offline_2020,
	title = {Offline reinforcement learning: {Tutorial}, review, and perspectives on open problems},
	journal = {arXiv preprint arXiv:2005.01643},
	author = {Levine, Sergey and Kumar, Aviral and Tucker, George and Fu, Justin},
	year = {2020},
}

@inproceedings{fujimoto2019off,
  title={Off-policy deep reinforcement learning without exploration},
  author={Fujimoto, Scott and Meger, David and Precup, Doina},
  booktitle={International conference on machine learning},
  pages={2052--2062},
  year={2019},
  organization={PMLR}
}

@article{kumar_conservative_2020,
	title = {Conservative $Q$-learning for offline reinforcement learning},
	volume = {33},
	journal = {Advances in Neural Information Processing Systems},
	author = {Kumar, Aviral and Zhou, Aurick and Tucker, George and Levine, Sergey},
	year = {2020},
	pages = {1179--1191},
}

@article{saxena_reinforcement_2021,
	title = {Reinforcement learning for efficient and tuning-free link adaptation},
	volume = {21},
	issn = {1536-1276},
	number = {2},
	journal = {IEEE Transactions on Wireless Communications},
	author = {Saxena, Vidit and Tullberg, Hugo and Jaldén, Joakim},
	year = {2021},
	note = {Publisher: IEEE},
	pages = {768--780},
}

@inproceedings{zuo2020transformer,
  title={Transformer hawkes process},
  author={Zuo, Simiao and Jiang, Haoming and Li, Zichong and Zhao, Tuo and Zha, Hongyuan},
  booktitle={International conference on machine learning},
  pages={11692--11702},
  year={2020},
  organization={PMLR}
}

@inproceedings{zhang2023continuous,
  title={Continuous-Time decision transformer for healthcare applications},
  author={Zhang, Zhiyue and Mei, Hongyuan and Xu, Yanxun},
  booktitle={International Conference on Artificial Intelligence and Statistics},
  pages={6245--6262},
  year={2023},
  organization={PMLR}
}

@article{chen_decision_2021,
	title = {Decision transformer: {Reinforcement} learning via sequence modeling},
	volume = {34},
	journal = {Advances in neural information processing systems},
	author = {Chen, Lili and Lu, Kevin and Rajeswaran, Aravind and Lee, Kimin and Grover, Aditya and Laskin, Misha and Abbeel, Pieter and Srinivas, Aravind and Mordatch, Igor},
	year = {2021},
	pages = {15084--15097},
}

@article{radford_language_2019,
	title = {Language models are unsupervised multitask learners},
	volume = {1},
	number = {8},
	journal = {OpenAI blog},
	author = {Radford, Alec and Wu, Jeffrey and Child, Rewon and Luan, David and Amodei, Dario and Sutskever, Ilya},
	year = {2019},
	pages = {9},
}

@article{fujimoto_benchmarking_2019,
	title = {Benchmarking batch deep reinforcement learning algorithms},
	journal = {arXiv preprint arXiv:1910.01708},
	author = {Fujimoto, Scott and Conti, Edoardo and Ghavamzadeh, Mohammad and Pineau, Joelle},
	year = {2019},
}

@article{emmons_rvs_2021,
	title = {Rvs: {What} is essential for offline {RL} via supervised learning?},
	journal = {arXiv preprint arXiv:2112.10751},
	author = {Emmons, Scott and Eysenbach, Benjamin and Kostrikov, Ilya and Levine, Sergey},
	year = {2021},
}

@inproceedings{schweighofer2022dataset,
  title={A dataset perspective on offline reinforcement learning},
  author={Schweighofer, Kajetan and Dinu, Marius-constantin and Radler, Andreas and Hofmarcher, Markus and Patil, Vihang Prakash and Bitto-Nemling, Angela and Eghbal-zadeh, Hamid and Hochreiter, Sepp},
  booktitle={Conference on Lifelong Learning Agents},
  pages={470--517},
  year={2022},
  organization={PMLR}
}

@misc{3rd_generation_partnership_project_3gpp_technical_2024,
	title = {Technical {Specification} ({TS}) 38.214. {NR}, {NR}; {Physical} layer procedures for data, v18.3.0},
	author = {3GPP},
	month = jun,
	year = {2024},
}

@article{levine_offline_2020-1,
	title = {Offline reinforcement learning: {Tutorial}, review, and perspectives on open problems},
	journal = {arXiv preprint arXiv:2005.01643},
	author = {Levine, Sergey and Kumar, Aviral and Tucker, George and Fu, Justin},
	year = {2020},
}

@ARTICLE{Ramireddy:22,
  author={Ramireddy, Venkatesh and Grossmann, Marcus and Landmann, Markus and Galdo, Giovanni Del},
  journal={IEEE Communications Standards Magazine}, 
  title={Enhancements on {Type-II 5G New Radio} Codebooks for {UE} Mobility Scenarios}, 
  year={2022},
  volume={6},
  number={1},
  pages={35-40},
  doi={10.1109/MCOMSTD.0001.2100072}
}

@article{Kumar:23,
  title={Offline $Q$-Learning on Diverse Multi-Task Data Both Scales And Generalizes},
  author={A. Kumar and R. Agarwal and X. Geng and G. Tucker and S. Levine},
  journal={International Conference on Learning Representations (ICLR)},
  year={2022},
}

@INPROCEEDINGS{Nicolini:23,
  author={Nicolini, Andrea and Icolari, Vincenzo and Dardari, Davide},
  booktitle={{IEEE} International Conference on Communications}, 
  title={Link Adaptation Algorithm for Optimal Modulation and Coding Selection in {5G} and Beyond Systems}, 
  year={2023},
  volume={},
  number={},
  pages={5279-5284},
  doi={10.1109/ICC45041.2023.10279293}
}

@ARTICLE{Chen:23,
  author={Chen, Jie and Ma, Juntao and He, Yihao and Wu, Gang},
  journal={IEEE Communications Letters}, 
  title={Deployment-Friendly Link Adaptation in Wireless Local-Area Network Based on On-Line Reinforcement Learning}, 
  year={2023},
  volume={27},
  number={12},
  pages={3424-3428},
  doi={10.1109/LCOMM.2023.3327964}
}

@article{lyu2022mildly,
  title={Mildly conservative $Q$-learning for offline reinforcement learning},
  author={Lyu, Jiafei and Ma, Xiaoteng and Li, Xiu and Lu, Zongqing},
  journal={Advances in Neural Information Processing Systems},
  volume={35},
  pages={1711--1724},
  year={2022}
}

@ARTICLE{Blanquez-Casado:19,
  author={Blanquez-Casado, Francisco and Aguayo Torres, Maria del Carmen and Gomez, Gerardo},
  journal={IEEE Communications Letters}, 
  title={Link Adaptation Mechanisms Based on Logistic Regression Modeling}, 
  year={2019},
  volume={23},
  number={5},
  pages={942-945},
  doi={10.1109/LCOMM.2019.2903045}
}

@INPROCEEDINGS{Bruno:14,
  author={Bruno, Raffaele and Masaracchia, Antonino and Passarella, Andrea},
  booktitle={2014 IEEE 80th Vehicular Technology Conference (VTC2014-Fall)}, 
  title={Robust Adaptive Modulation and Coding {(AMC)} Selection in {LTE} Systems Using Reinforcement Learning}, 
  year={2014},
  volume={},
  number={},
  pages={1-6},
  doi={10.1109/VTCFall.2014.6966162}
}

@book{Sutton:14,
  title={Reinforcement Learning: An Introduction (Second Edition)},
  author={Richard S. Sutton and Andrew G. Barto},
  year={2014},
  publisher={{MIT Press}},
}

@misc{Peri:24,
      title={Offline Reinforcement Learning and Sequence Modeling for Downlink Link Adaptation}, 
      author={Samuele Peri and Alessio Russo and Gabor Fodor and Pablo Soldati},
      year={2024},
      eprint={2410.23031},
      archivePrefix={arXiv},
      url={https://arxiv.org/abs/2410.23031}, 
}
\ifdefined\addappendix

\onecolumn
\appendix
\subsection{Hyperparameters and simulation parameters}
Table \ref{table: hyperparameters} reports the hyperparameters chosen for the three offline \ac{RL} methods we considered in this work. 

\begin{table}[h]
    \centering
    \begin{minipage}{0.48\textwidth}
        \centering
        \caption{Hyperparameters for the offline RL methods.}
        \label{table: hyperparameters}
        \begin{tabular}{lM}
            \toprule
            \multicolumn{2}{l}{\textbf{CQL Parameters}} \\
            Value-network \#layers & 3 \text{ ($\mathcal{D}_\text{opt}$)} \ | \  7 \text{ ($\mathcal{D}_\text{s-opt}$)} \\
            Value-network act. function & \text{ReLU activation} \\
            Training steps & 100000 \\
            Learning rate & 5 \cdot 10^{-4} \\
            Batch size & 64 \\
            Target update interval & 2000 \\
            $\alpha$ & 0.2 \\
            \midrule
            \multicolumn{2}{l}{\textbf{BCQ Parameters}} \\
            Value-network \#layers & 3 \\
            Value-network act. function & \text{ReLU} \\
            Training steps & 100000 \\
            Learning rate & 1 \cdot 10^{-4} \\
            Batch size & 64 \\
            Target update interval & 1024 \\
            $\tau$ & 0.5 \\
            $\beta$ & 0.3 \\
            \midrule 
            \multicolumn{2}{l}{\textbf{DT Parameters}} \\
            Number of attn. layers & 4 \\
            Number of attn. heads & 8 \\
            Embedding dimension ($d_\text{model}$) & 256 \\
            Activation function & \text{GeLU} \\
            Batch size & 64 \\
            Context length ($K$) & \min(32, 5n_p) \\
            Training steps & 150000 \\
            Learning rate & 3 \cdot 10^{-6} \\
            Optimizer type & \text{Adam} \\
            Weight decay & 1 \cdot 10^{-4} \\
            \bottomrule
        \end{tabular}
    \end{minipage}%
    \hfill
\end{table}

\subsection{Additional DT conditioning results.}
\label{Appendix: DT}
In this appendix, we provide additional results supporting our claim that the stochasticity intrinsic to the \ac{LA} process prevents \ac{DT} from correctly conditioning the inference of optimal actions when considering long sequences of transitions.
\\
\subsubsection*{\textbf{Sequences of $H$ transitions}}
We repeat the same analysis conducted in Section \ref{RTG conditioning} by choosing a fixed number of transitions to include inside input sequences for \ac{DT}, regardless of the number of packets being considered.

Table \ref{table:vanilla sweep} shows the performance in terms of throughput, \ac{BLER} and \ac{SE} achieved by the same \ac{DT} model conditioned on increasing target \ac{RTG} values $\omega$ over the $q\%$ quantiles of the training distribution of \acp{RTG} (with $q \in [0.1, 0.2, ..., 1.0]$). The model takes input sequences of $H=32$ transitions. Similarly to what we observed in Section \ref{RTG conditioning}, the effect of conditioning appears negligible on each of the three considered metrics.

In Table \ref{table:3gpp sweep}, similarly to \ac{3GPP}, we condition action generation on a different target depending on the \ac{CQI} estimate included in the state $s_t$. Instead of referring to the \ac{CQI} Table 5.2.2.1-3 of the \ac{3GPP} specification~\cite{3rd_generation_partnership_project_3gpp_technical_2024}, we choose to condition, for each packet, over the $q\%$ quantiles of the training distribution of \acp{RTG} of packets with equal \ac{CQI} to the current one. Since, also in this case, there is no correlation between the performance and the conditioning variable, the correlation between achievable rewards and channel conditions is not the only reason behind the issues we faced in correctly conditioning the inference of actions.\\

\begin{table}[ht]
\centering
\begin{minipage}{0.48\linewidth}
\centering
\captionsetup{width=0.9\linewidth}
\caption{Performance obtained by conditioning on increasing, fixed, RTG values at inference (vanilla).}
\label{table:vanilla sweep}
\begin{tabular}{cccc}
\toprule\toprule
$\omega$           & Throughput       & BLER            & SE               \\ \hline
1.18       &  15.90 Mbps & 0.35           & 7.29 bps/Hz \\
1.71       &   15.90 Mbps & 0.35          & 7.30 bps/Hz \\
2.18       &   15.91 Mbps & 0.35          & 7.30 bps/Hz \\
2.54       &    15.90 Mbps & 0.35         & 7.30 bps/Hz \\
3.03       & 16.03 Mbps    & 0.30         & 7.44 bps/Hz  \\
3.86       & 16.04 Mbps    & 0.31         & 7.45 bps/Hz  \\
4.19       & 16.01 Mbps    & 0.31         & 7.45 bps/Hz  \\
5.30       & 16.01 Mbps    & 0.31         & 7.46 bps/Hz  \\
5.92       & 16.01 Mbps    & 0.31         & 7.47 bps/Hz  \\
7.57       & 15.99 Mbps    & 0.32         & 7.47 bps/Hz  \\
\bottomrule\bottomrule
\end{tabular}
\end{minipage}%
\begin{minipage}{0.48\linewidth}
\centering
\captionsetup{width=0.9\linewidth}
\caption{Performance obtained by conditioning on increasing RTG values (per CQI) at inference (3GPP).}
\label{table:3gpp sweep}
\begin{tabular}{cccc}
\toprule\toprule
 $q$          & Throughput       & BLER            & SE               \\ \hline
0.1       & 16.03 Mbps & 0.30    & 7.43 bps/Hz          \\
0.2       & 16.05 Mbps & 0.30 & 7.44 bps/Hz             \\
0.3       & 16.05 Mbps & 0.30       & 7.44  bps/Hz    \\
0.4       & 16.07 Mbps & 0.30        & 7.44 bps/Hz  \\
0.5       & 16.04 Mbps    & 0.31       & 7.44 bps/Hz  \\
0.6       & 15.97 Mbps    & 0.31       & 7.43 bps/Hz  \\
0.7       & 16.03 Mbps    & 0.31       & 7.44 bps/Hz  \\
0.8       & 16.04 Mbps    & 0.31        & 7.45 bps/Hz  \\
0.9       & 16.05 Mbps    & 0.31     & 7.45 bps/Hz  \\ 
\bottomrule\bottomrule
\end{tabular}
\end{minipage}
\end{table}

\subsubsection*{\textbf{Influence of stochasticity in a simple environment}}
We found it convenient to design a much-simplified model of the \ac{LA} environment to better investigate its properties with relation to the \ac{RL} models we considered. The environment maintains the main features of the process simulated by the internal RAN simulator.

\begin{figure}
    \centering
    \includegraphics[width=0.6\textwidth]{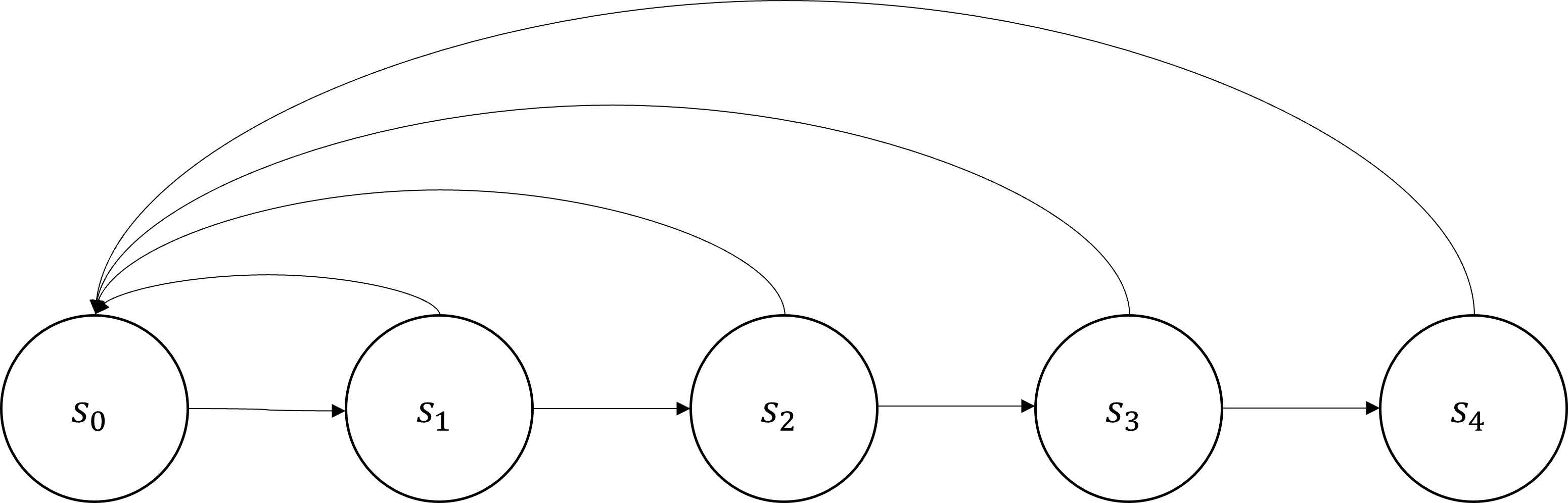}
    \caption{Custom LA environment.}
    \label{fig:custom environment}
\end{figure}

As shown in figure \ref{fig:custom environment}, the environment consists of $n=5$ discrete states, each representing the $i^{\text{th}}$ transmission attempt of a packet, with $i \in [0,n)$. The action space $\mathcal{A}$ is the same of the original link adaptation process, comprising 28 actions represented, this time, by integers in $[1,28]$. Higher indices represent actions that are more aggressive and are, therefore, more likely to cause a transmission to fail. Failing transmissions in state $i$ makes the agent transition to either state $i+1$ or, if in $s_4$, to the initial state $s_0$. Successful transmissions bring the agent back to the initial state $s_0$.

The environment dynamics are function of a context $x \in [0,500)$ representing the quality of the downlink channel, context values are sampled uniformly at random at each step of the \ac{MDP}. Given an action, high/low values of $x$ increase/decrease the probability of experiencing successful transmissions:
\begin{equation}
    p_{\text{success}} = p(s_{k+1}|s_k, a_t, x_t) = \tanh\left({\frac x {18 \cdot a_t}}\right).
\end{equation}
Aggressive actions are awarded higher rewards in case of success while penalties for failed transmissions increase with the number of transmission attempts:
\begin{equation}
     R(s_t,a_t,x_t)=\begin{cases}
        \tanh\left(\frac {a_t} {28}\right) & \text{if SUCCESS} \\
        -\beta (k+1) & \text{otherwise.}
        \end{cases}
\end{equation}
Having a simple version of the environment allows to compute exact values of states and actions through dynamic programming and, therefore, find out optimal actions. Figure~\ref{fig:q values} plots the q-values of state-action pairs relative to $s_0$ while Figure~\ref{fig:optimal actions} shows the frequency of each action being optimal for different values of context (both the plots refer to $s_0$).

\begin{figure}[ht]
    \centering
    \begin{minipage}{0.62\textwidth}
        \centering
        \includegraphics[width=\textwidth]{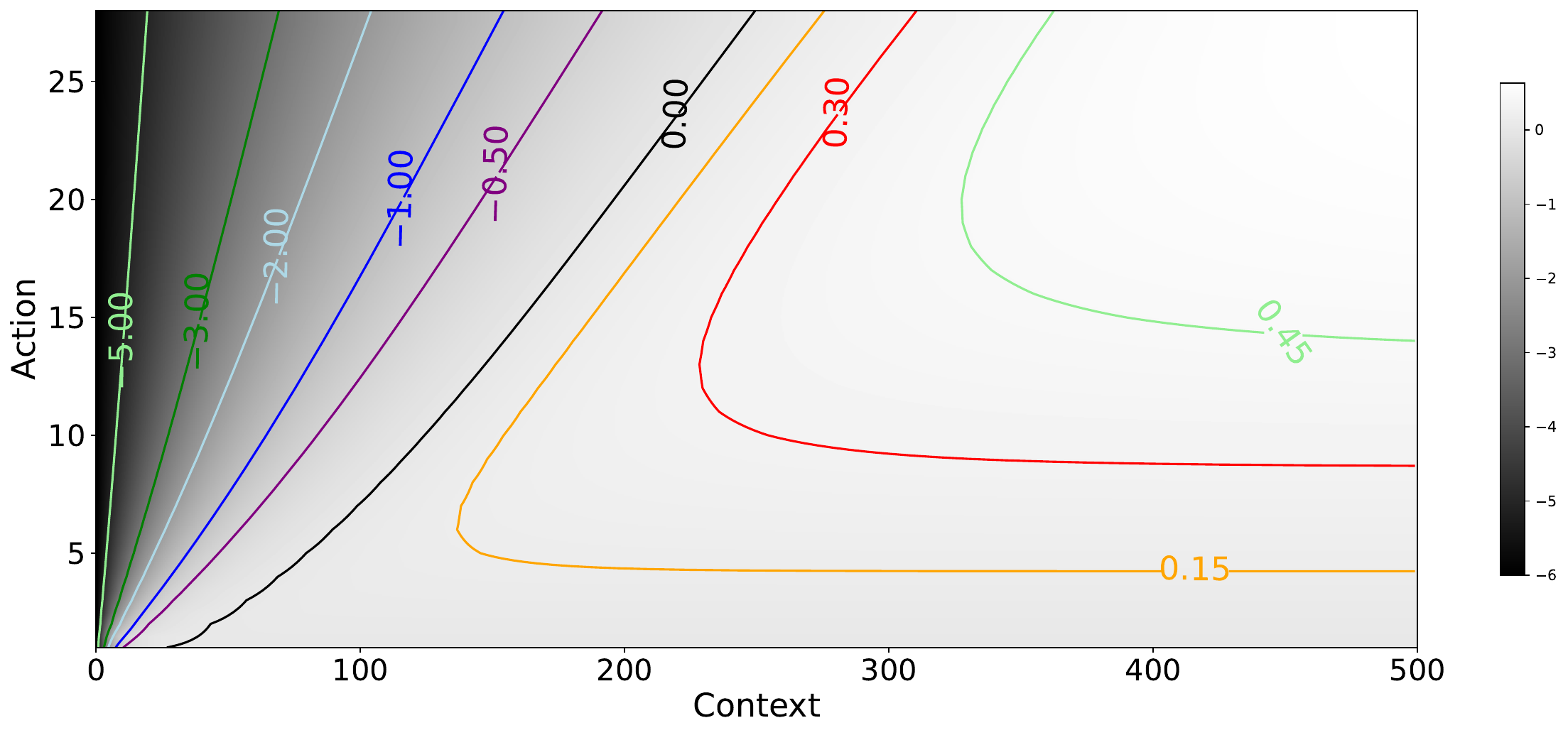}
        \caption{Q-values at $s_0$.}
        \label{fig:q values}
    \end{minipage}%
    \hfill
    \begin{minipage}{0.36\textwidth}
        \centering
        \includegraphics[width=\textwidth]{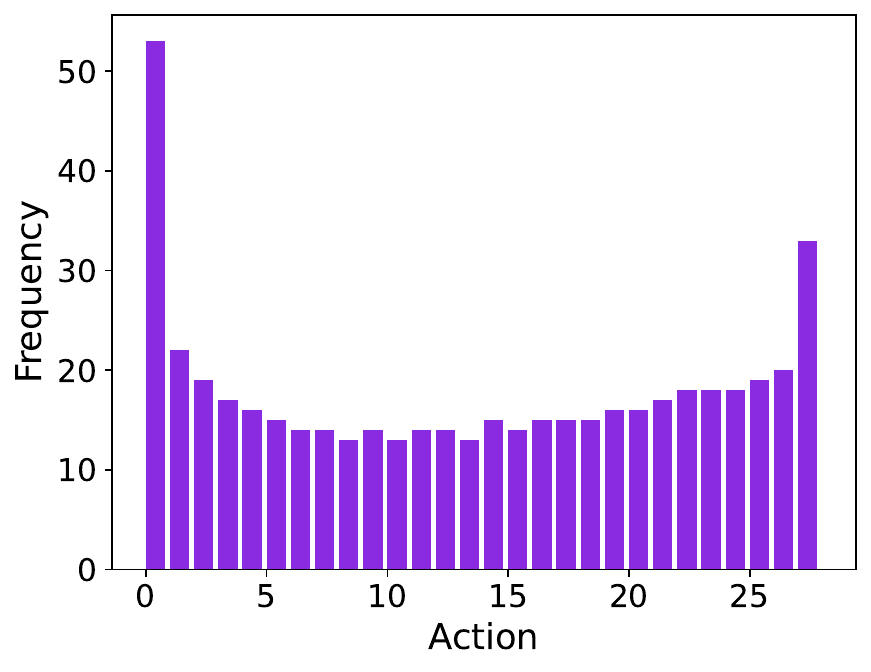}
        \caption{Frequency of each action being optimal in $s_0$.}
        \label{fig:optimal actions}
    \end{minipage}
\end{figure}

\section{Custom environment results}
\label{section:Custom environment results}
Through dynamic programming we compute the value of actions for each state-context pair and use the greedy policy to collect data with different levels of randomization $\epsilon$. We then train and test \ac{DT} agents over 5 different seeds and show average, total returns and relative standard deviation obtained conditioning on the 0.1, 0.3, 0.5, 0.7 and 0.9 quantiles of the RTG distribution inside the dataset (the one collected without taking random actions). Table \ref{table: custom-env-results} shows how, even on our custom environment, conditioning on different fixed values at inference has no effect. These results support our hypothesis that the various sources of stochasticity inherent to \ac{LA} negatively impact the effectiveness of \ac{DT}.

\begin{table}[]
\centering
\caption{Total returns achieved by DT conditioned on five different values, compared against performance of the Greedy agent on our custom environment.}
\renewcommand{\arraystretch}{1.5}
\begin{tabular}{lllll}
\toprule\toprule
Method       & $\epsilon=0.0$           & $\epsilon=0.25$          & $\epsilon=0.5$           & Avg \\ \hline
Greedy agent &   12306.35 $\pm$ 39.82 &  10820.35 $\pm$ 43.33 & 9302.41 $\pm$ 58.99&   10809.70  \\ \hline
DT(10\%)     & 5275.12 $\pm$ 505.22           & 5571.18 $\pm$ 175.69             & 4859.03 $\pm$ 285.87            &   5235.11  \\
DT(30\%)     & 5280.40 $\pm$  517.79           & 5591.95 $\pm$ 161.22            & 4791.11 $\pm$ 249.83            &  5221.16   \\
DT(50\%)     & 5314.01$\pm$ 566.83             & 5574.35 $\pm$ 203.88            & 4827.60 $\pm$ 237.40            &   5238.65  \\
DT(70\%)     & 5278.47 $\pm$ 520.93            & 5568.78 $\pm$ 167.02            & 4819.47 $\pm$ 279.05            &  5222.22   \\
DT(90\%)     & 5271.64  $\pm$ 504.16           & 5552.28 $\pm$ 187.44             & 4813.07 $\pm$ 305.59            &  5212.33   \\
\bottomrule\bottomrule
\end{tabular}

\label{table: custom-env-results}
\end{table}

\label{sec:appendix}
\else\fi
\end{document}